\documentclass[sigconf,nonacm]{acmart}

\AtBeginDocument{%
  \providecommand\BibTeX{{%
    \normalfont B\kern-0.5em{\scshape i\kern-0.25em b}\kern-0.8em\TeX}}}

\usepackage[utf8]{inputenc} 
\usepackage[T1]{fontenc}    
\usepackage{hyperref}       
\usepackage{url}            
\usepackage{booktabs}       
\usepackage{tabularx}
\usepackage{nicefrac}       
\usepackage{microtype}      
\usepackage{graphicx}
\usepackage{subfigure}
\usepackage{wrapfig}
\usepackage{lipsum}
\usepackage{multirow}
\usepackage[listings,skins,breakable]{tcolorbox}
\usepackage{algorithm}      
\usepackage{algorithmic}    
\usepackage{xcolor}
\usepackage{bbding}         
\usepackage{duckuments}     

\usepackage{CJKutf8} 

\usepackage[capitalize,noabbrev]{cleveref}

\theoremstyle{plain}
\newtheorem{theorem}{Theorem}[section]

\newtheorem{lemma}[theorem]{Lemma}

\theoremstyle{definition}

\theoremstyle{remark}

\usepackage[textsize=tiny]{todonotes}

\begin{document}

\title{Contrastive Private Data Synthesis via Weighted Multi-PLM Fusion}

\author{Tianyuan Zou}
\affiliation{%
  \institution{Institute for AI Industry Research (AIR), Tsinghua University, Beijing}
  \country{China}
}

\author{Yang Liu}
\authornote{Correspondence: yangliu62@tsinghua.edu.cn}
\affiliation{%
  \institution{Institute for AI Industry Research (AIR), Tsinghua University, Beijing} 
  \country{China}
}

\author{Peng Li}
\affiliation{%
  \institution{Institute for AI Industry Research (AIR), Tsinghua University, Beijing}
  \country{China}
}

\author{Yufei Xiong}
\affiliation{%
 \institution{the Department of Mathematics, Harbin Institute of Technology, Weihai, Shandong}
 \country{China}
 }

\author{Jianqing Zhang}
\affiliation{%
  \institution{Shanghai Jiao Tong University, Shanghai}
  \country{China}
}

\author{Jingjing Liu}
\affiliation{%
  \institution{Institute for AI Industry Research (AIR), Tsinghua University, Beijing}
  \country{China}
}

\author{Xiaozhou Ye and Ye Ouyang}
\affiliation{%
  \institution{AsiaInfo Technologies, Shanghai}
  \country{China}
  }

\author{Ya-Qin Zhang}
\affiliation{%
  \institution{Institute for AI Industry Research (AIR), Tsinghua University, Beijing}
  \country{China}
  }

\renewcommand{\shortauthors}{Zou, et al.}

\begin{abstract}
Substantial quantity and high quality are the golden rules of making a good training dataset with sample privacy protection equally important. 
Generating synthetic samples that resemble high-quality private data while ensuring Differential Privacy (DP), a formal privacy guarantee, promises scalability and practicality.
However, existing methods relying on pre-trained models for data synthesis 
often struggle in data-deficient scenarios, suffering from limited sample size, inevitable generation noise and existing pre-trained model bias. 
To address these challenges, we propose a novel contr\textbf{A}stive private data \textbf{S}ynthesis via \textbf{W}eighted multiple \textbf{P}re-trained language models (PLM) framework, named as \textbf{WASP}. 
WASP utilizes limited private samples for more accurate private data distribution estimation via a Top-$Q$ voting mechanism, and leverages low-quality synthetic samples for contrastive generation via collaboration among dynamically weighted multiple pre-trained models.
Extensive experiments on $6$ well-developed datasets with $6$ open-source and $3$ closed-source PLMs demonstrate the superiority of WASP in improving model performance over diverse downstream tasks.
Code is available at \url{https://anonymous.4open.science/r/WASP}. \end{abstract}





\received{31 January 2025}
\received[revised]{31 January 2025}
\received[accepted]{31 January 2025}

\maketitle

\section{Introduction}
In the rapidly evolving landscape of AI models and AI agents, the strength of both Large Language Models (LLMs) and Small Task-specific Models (STMs) 
hinges on the abundance of high-quality training data~\cite{budach2022effects,wang2024survey}, of which only a limited amount of samples can be harnessed in practice. To further complicate the issue, broad tasks across disciplines such as medical record summarization~\cite{rumshisky2016predicting},  chatbots for personalized weight loss~\cite{chew2022use} and instruction-following LLM fine-tuning~\cite{yu2024privacypreserving} all rely on high-quality private data collected from real users, which inevitably incurs non-negligible privacy issues.

Differentially private synthetic data stands in as a promising remedy~\cite{bommasani2019towards,putta2023differentially,flemings2024differentially}, by creating a new synthetic dataset that resembles the real private dataset while preserving the privacy of each sample via guaranteeing Differential Privacy (DP)~\cite{dwork2006differential}. 
There are two main lines of research for generating DP synthetic datasets.
The first line of works~\cite{mattern2022differentially,yue2023synthetic} introduce DP Fine-tune Generator which 
involves fine-tuning a Pre-trained Language Model (PLM) using DP-SGD~\cite{abadi2016deep}. 
However, this practice is computationally intensive and requires substantial data for effective fine-tuning.
Another line of work, Private Evolution (PE)~\cite{lin2024differentially,xie2024differentially,hou2024pretext}, relieves the fine-tuning requirement and instead merely uses the APIs of pre-trained models for generation, under DP-protected guidance from private samples. This API-based nature is efficient in creating DP synthetic data, and can leverage both open-source and closed-source pre-trained models, making PE a more applicable solution compared to its counterparts.

\begin{figure*}
    \centering
    \subfigure[]{
         \begin{minipage}[t]{0.48\linewidth}
         \centering
         \includegraphics[width=1\linewidth]{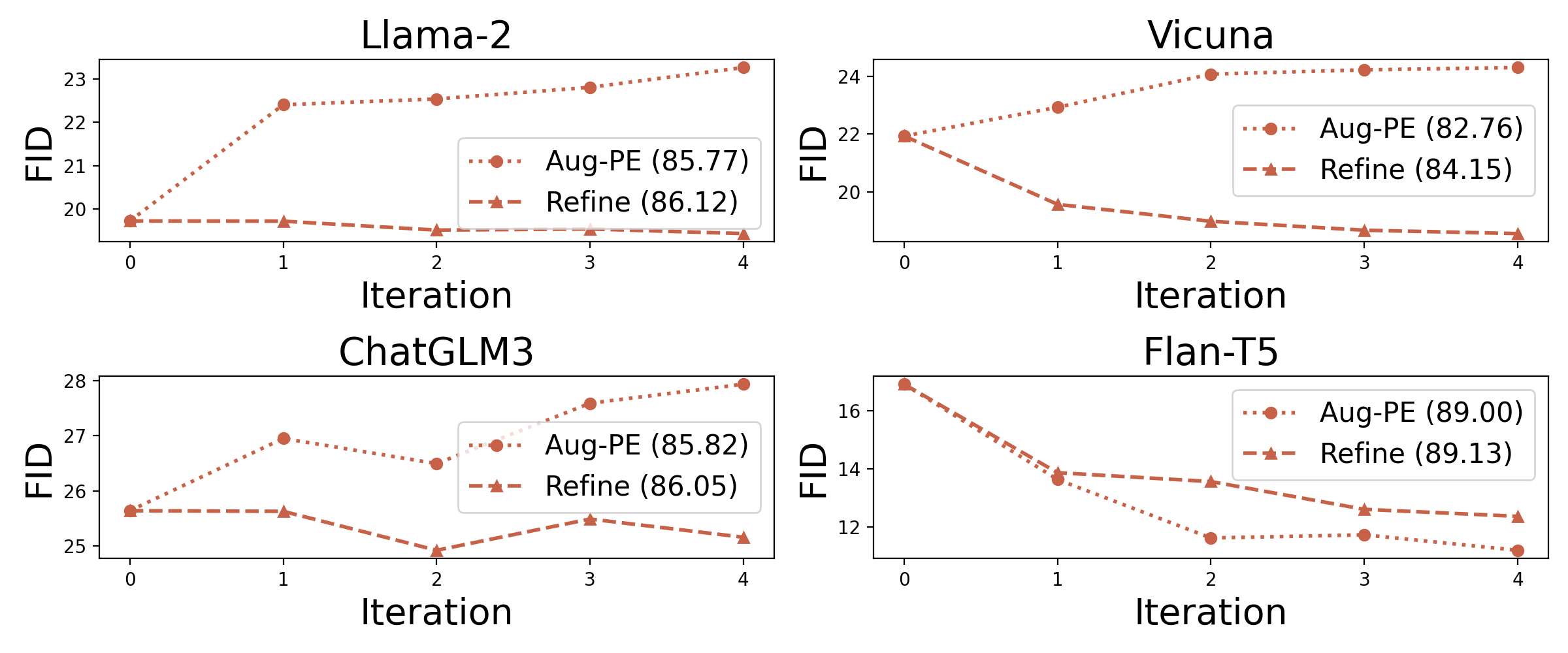}
         \vspace{-1em}
         \label{subfig:pe_random_fid_and_acc}
         \end{minipage}
     }
    \subfigure[]{
         \begin{minipage}[t]{0.48\linewidth}
         \centering
         \includegraphics[width=1\linewidth]{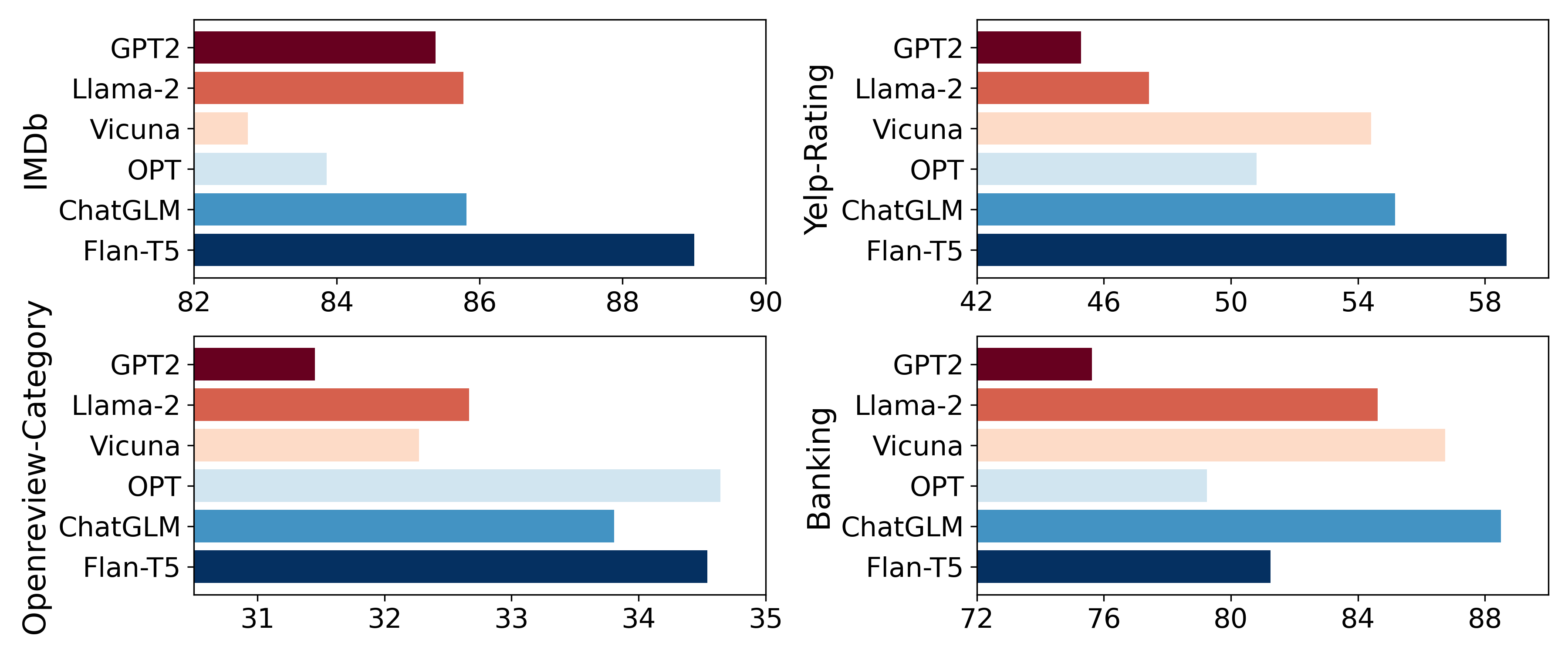}
         \vspace{-1em}
         \label{subfig:pe_variance}
         \end{minipage}
     }
     \vspace{-0.7em}
     \caption{$(a)$ Comparison of the similarity of synthetic dataset to real private dataset (measured by FID~\cite{heusel2017gans}) and STM performance (numbers within parenthesis) of Aug-PE~\cite{xie2024differentially} (dotted lines) and our refinement (dashed lines) under $(4.0, 1\times10^{-5})$-DP with IMDb dataset. Lower FID indicates higher similarity. $(b)$ Results of Aug-PE using $100$ private samples and $(4.0, 1\times10^{-5})$-DP.}
    \label{fig:intro_fig_1}
    \vspace{-0.5em}
\end{figure*}

Although proven effective, current PE methods~\cite{lin2024differentially,xie2024differentially,hou2024pretext}, 
 still face three major challenges:
\textbf{$(1)$ Limited Private Samples.}
Existing PE methods rely on at least thousands of private samples~\cite{lin2024differentially,xie2024differentially,hou2024pretext} to guarantee reliable generation feedback selection.
In practice, however, data sources may provide only a few hundred samples~\cite{zdrazil2024chembl,ren2019almost}, leading to noisy selection guidance.
As shown in \cref{subfig:pe_random_fid_and_acc}, with limited private samples ($100$), Aug-PE~\cite{xie2024differentially} (PE for text)
failed to generate synthetic samples resembling real samples' distribution
for $3$ PLMs (except Flan-T5).
Similar conclusion is drawn  in~\citet{lin2024differentially} (see Table 2 therein).
This calls for a more precise guidance from limited private samples.
\textbf{$(2)$ Noisy Synthetic Data.}
Although PE approaches encourage the generation of high-quality samples that are close to real private sample distribution, low-quality noisy samples are still unavoidable (see examples in \cref{tab:examples_bad_samples} in \cref{subsec:appendix_good_bad_samples}),  
which hinder the final performance when training downstream models~\cite{ye2022progen,gao2023self,zou2024fusegen}. This highlights the the importance of instructing the avoidance of generating noisy samples during data synthesis.
\textbf{$(3)$ Risky PLM Selection.} 
As shown in \cref{subfig:pe_random_fid_and_acc},
different PLMs yield varying performances (some with unsatisfactory results), and even the best performing model differs across various downstream tasks (see \cref{subfig:pe_variance}), making it hard to select the most suitable pre-trained model for a specific task. Previous PE works primarily focus on single PLM setting, thus the potential of collaboration among multiple PLMs is still unexplored.
To address these demanding challenges, we propose WASP, a collaborative framework that fuses the knowledge from weighted multiple PLMs to synthesize DP data in a contrastive fashion. 
$(1)$ \textit{To overcome private sample scarcity}, we
first extend the voting mechanism for private distribution estimation used in PE from Top-$1$ to Top-$Q$ ($Q>1$) with decaying weights, 
in order to get a more accurate estimation while guaranteeing private data DP. 
$(2)$ \textit{To reduce noise}, we
further leverage the previous voting results to select both high-quality and low-quality samples, and incorporate a contrastive prompt containing both types of samples to improve synthetic data quality by encouraging generation that is more aligned to high-quality samples and less similar to low-quality ones.
Notice that under multi-PLM setting, these samples may originate from different PLMs. 
$(3)$ \textit{To mitigate model bias}, we then interfuse the capabilities of multiple PLMs with dynamically
adjusted importance weight for each PLM based on the ensemble votes from private samples. 
The underlying principle is to assign higher weights to PLMs that generate synthetic samples that are closer to real samples on average. Operating in an iterative fashion, the WASP framework can generate large quantity of synthetic data that better approximate real private data distribution while observing differential privacy. Notably, this process incurs no additional API queries compared to its single-PLM counterparts.

Our contributions are summarized as follows:

$1)$ We introduce a privacy-preserving collaborative framework WASP to facilitate collaboration between multiple PLMs and private samples, especially benefiting scenarios with limited private data.

$2)$ Our proposed WASP leverages differentially private Top-$Q$ voting to improve the estimation of private distributions using limited private samples. It generates higher-quality data by contrasting high- and low-quality samples and dynamically assigns importance weights to PLMs, ensuring that more capable PLMs of the specific task are prioritized. 

$3)$ Experiments on $6$ well-defined natural language processing tasks with $6$ open-source and $3$ closed-source PLMs demonstrate the consistent superiority of WASP over existing methods,
especially for challenging tasks.

\begin{figure*}[!tb]
    \centering
    \includegraphics[width=0.99\linewidth]{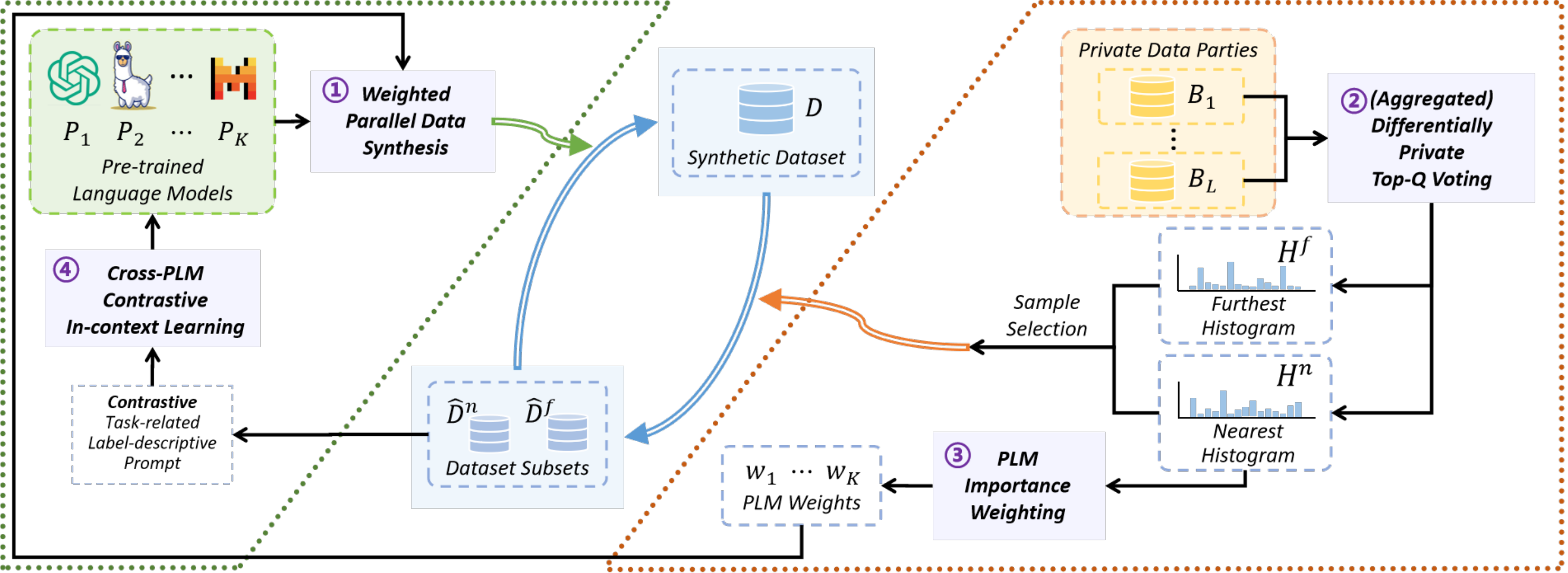}
    \caption{Overview of WASP framework.} 
    \label{fig:framework}
\end{figure*}

\section{Related Work}


\textbf{DP Synthetic Dataset.} 
The goal of generating DP synthetic data is to mimic private dataset while protecting sensitive information. 
Although fine-tuning a PLM with DP-SGD~\cite{abadi2016deep} for data generation purpose can be effective
~\cite{bommasani2019towards,putta2023differentially,flemings2024differentially,mattern2022differentially,yue2023synthetic}, it 
is computationally intensive and requires a large number of high-quality private samples to reach strong performance.
Moreover, many state-of-the-art PLMs such as GPT series~\cite{openai2021gpt3-5,openai2023gpt4,hurst2024gpt4o} are also closed-source, making DP fine-tuning impractical.

A new line of work instead relies on generative APIs of PLMs without fine-tuning, which focuses on either iterative data synthesis under DP guidance~\cite{lin2024differentially,zhao2024generate,bojkovic2024differentially} or creating DP replica of a given large private dataset~\cite{nagesh2024private}. 
Given that requiring a large global dataset for synthetic data initialization~\cite{zhao2024generate} is hard to obtain in most cases, \citet{lin2024differentially} proposes a more practical solution, Private Evolution (PE), which instead uses task-related synthetic samples. In PE, private samples are used to identify their nearest synthetic counterparts under DP protection, which then guide the growth of the DP synthetic dataset. PE is proven effective across images~\cite{lin2024differentially} and text~\cite{xie2024differentially}, and is further adapted to federated private data scenarios~\cite{hou2024pretext}.
However, all these works primarily focus on using a single PLM as the generation model.

\textbf{PLM Fusion.} The combination of multiple PLMs can lead to stronger model performance~\cite{liu2023dynamic,du2023improving,wan2024knowledge,wan2024fusechat,li2024more,zou2024fusegen}. Some studies fine-tune target models with token-level fusion from PLMs as teachers during training time~\cite{wan2024knowledge,wan2024fusechat}, while others use majority voting~\cite{li2024more} or logits averaging~\cite{mavromatis2024pack} for knowledge fusion during inference. 
However, data privacy challenges persist, as training or test samples are exposed to external models. To solve this, FuseGen~\cite{zou2024fusegen} recently proposes PLM fusion in a zero-shot learning setting, utilizing only model APIs to synthesize data without accessing real private samples, thereby ensuring data privacy. However, by treating all PLMs equally, it overlooks the capability difference of individual PLMs over different tasks.
More related works considering Contrastive In-context Learning are included in \cref{sec:appendix_related_work}.


\section{Preliminaries}
\textbf{Differential Privacy (DP).} 
If two datasets $\mathcal{D}$ and $\mathcal{D}'$ differ in a single entry, 
they are referred to as \textit{Neighboring Datasets}. 
A mechanism $\mathcal{M}$ satisfies $(\epsilon,\delta)$-DP if for any neighboring datasets $\mathcal{D},\mathcal{D}'$ and any output subset $E$ of $\mathcal{M}$, it holds that~\cite{dwork2006differential}:
\begin{equation}
\text{Pr}[\mathcal{M}(\mathcal{D})\in E] \leq e^{\epsilon}\cdot\text{Pr}[\mathcal{M}(\mathcal{D}')\in E]+\delta. 
\end{equation}
Note that post-processing on the output of $(\epsilon,\delta)$-DP does not incur additional  privacy loss~\cite{dwork2014algorithmic}.

\textbf{Gaussian Mechanism.}
\textit{Gaussian Mechanism}~\cite{dwork2006differential} can be applied to guarantee $(\epsilon,\delta)$-DP, for any $\epsilon>0,\delta\in(0,1)$, by adding Gaussian noise following $\mathcal{N}(0,\sigma^2)$ to the transmitted statistics with $\sigma=\Delta\frac{\sqrt{2\log\left({1.25/\delta}\right)}}{\epsilon}$ and $\Delta$ being the sensitivity of $\mathcal{M}$~\cite{balle2018improving}.

\section{Methodology}

\subsection{Problem Definition}

In this paper, we aim to generate a DP synthetic dataset $\mathcal{D}=\{(\mathbf{x}_{i},y_{i})\}_{i=1}^{N}$ of size $N$ using a small number of private data $\mathcal{B}=\{(\mathbf{z}_{j},u_{j})\}_{j=1}^{M}$, where $M$ denotes the number of private samples, and $\mathbf{z}_{j},u_j$ denote the  feature and label of the private sample $j$, respectively. We consider  data-scarcity setting where $M$ is typically a few hundreds at most. To achieve this, we harness the collaborative power of $K$ black-box PLMs $\{\mathcal{P}_k\}_{k=1}^K$ via APIs, while protecting  private data by Gaussian DP.  
For evaluation, we use $\mathcal{D}$ to train a Small Task-specific Model (STM) $m$ and evaluate model performance on a test dataset $\mathcal{A}$ containing real samples that is never used during training. 

Note that our framework can be easily extended to the scenario of distributed federated data where each 
data source possesses an insufficient amount of private data and collaborates on private tasks with secure aggregation~\cite{hou2024pretext}. We present the related details
in \cref{subsec:multi_pdp}.

\subsection{Overall Workflow of WASP}
The overall workflow of WASP is depicted in \cref{fig:framework} and \cref{alg:algorithm_full_functions_singlePDP}, where four steps are taken iteratively for $T$ times. 
For a given task, the first iteration begins by prompting each PLM $\mathcal{P}_k$ with a zero-shot prompt, which describes the task and category label, to generate a synthetic data subset $\mathcal{D}_{k}$ of equal size $N_k$.
These samples do not contain information about $\mathcal{B}$. 
The collective dataset $\mathcal{D}=\bigcup_{k=1}^K\mathcal{D}_k$ is then voted by each private sample using a differentially private Top-$Q$ voting mechanism to identify high-quality and low-quality synthetic samples based on their similarity to the distribution of $\mathcal{B}$. These samples are then used to create a contrastive in-context learning prompt for the next round of PLM generation. The voting results are further exploit to dynamically adjust the importance weight $w_k$ for each PLM $\mathcal{P}_k$, which determines $N_k$ of the next generation round. The process repeats from here, expanding $\mathcal{D}$ with DP synthetic samples.
After $T$ iterations, $\mathcal{D}$ is used to train an STM $m$.
For notational simplicity, we omit the iteration index $t$, with $\mathcal{D}$ accumulated over iterations. 
DP guarantee of WASP is given in \cref{theorem:dp_L_equals_1} with proof included in \cref{sec:appendix_privacy_analysis}.
\begin{theorem} \label{theorem:dp_L_equals_1}
    WASP (\cref{alg:algorithm_full_functions_singlePDP}) satisfies $(\epsilon,\delta)$-DP.
\end{theorem}


\subsection{Weighted Parallel Data Synthesis} \label{subsec:method_step1}
In this stage (lines 4-6 in \cref{alg:algorithm_full_functions_singlePDP}), each PLM $\mathcal{P}_k$ generates $ N_k = \left[ (N/T)\times w_k \right]$ synthetic samples following:
\begin{equation} \label{eq:generation}
  \mathbf{x}_{i} \sim \mathcal{P}_k\left(\cdot|\mathcal{T}(y_{i})\right) \, ,
\end{equation}
where $\{w_k\}_{k=1}^K$ are the weights for $\{\mathcal{P}_k\}_{k=1}^K$, $\left[ \cdot \right]$ is the rounding function, $N$ is the expected total number of synthetic samples to be generated, and $\mathcal{T}(\cdot)$ is the 
generation prompt. In the initial iteration, $\mathcal{T}(\cdot)$ is a zero-shot prompt that describes the task and provides category description, with all PLMs receiving equal weights, i.e. $\{w_k=\frac{1}{K}\}_{k=1}^K$. For later iterations, $\mathcal{T}(\cdot)$ is extended to a few-shot contrastive prompt (see \cref{subsec:method_step4}) with in-context samples selected in \cref{subsec:method_step2}, and $\{w_k\}_{k=1}^K$ dynamically assigned based on each PLM’s capability of the specific task (see \cref{subsec:method_step3}).
The collective synthetic dataset $\mathcal{D}=\bigcup_{k=1}^K\mathcal{D}_k$ is then sent to the private data party.

\begin{algorithm}[tb]
\caption{WASP} 
\label{alg:algorithm_full_functions_singlePDP}

\begin{flushleft}
\textbf{Input:} \\
$K$ PLMs $\{\mathcal{P}_{k}\}_{k=1}^K$ with empty synthetic dataset $\{\mathcal{D}_k\leftarrow\emptyset\}_{k=1}^K$; 1 data party with private dataset $\mathcal{B}$ of size $M$ belonging to $C$ categories; 
number of in-context samples $S$;
number of iterations $T$ taken to obtain in total $N$ synthetic samples; 
initialized PLM weights ${\{w_{k}=1/K\}}_{k=1}^K$; 
learning rate $\eta$;
DP privacy parameters $\epsilon,\delta$;
training time unseen test dataset $\mathcal{A}$;
random initialized STM $m_{(0)}$.
\\
\textbf{Output:} STM $m$. 
\end{flushleft}

\begin{algorithmic}[1]
    \STATE Initialize in-context feedback samples $\hat{\mathcal{D}}^n \leftarrow \emptyset, \hat{\mathcal{D}}^f \leftarrow \emptyset$.
    \STATE Calculate Gaussian noise $\sigma=4\frac{\sqrt{2\log{\left(1.25/\delta\right)}}\sqrt{T-1}}{\epsilon}$.
    \FOR{$t=0$ {\bfseries to} $T-1$}
        \FOR{$k=1$ {\bfseries to} $K$ {\bfseries in parallel}}
            \STATE $\mathcal{D}_k \leftarrow$ \verb|WeightedSynDataGeneration(|$\mathcal{D}_k$, $\hat{\mathcal{D}}^n$, $\hat{\mathcal{D}}^f$, $\left[ (N/T)\times w_{k} \right]$, $C$\verb|)|.
        \ENDFOR
        \STATE $\mathcal{D} \leftarrow \cup_{k=1}^{K}\mathcal{D}_k$.
        \STATE $H^{n}, H^{f} \leftarrow$ \verb|DP_PrivateVoting(|$\mathcal{D}$, $\mathcal{B}$, $Q$, $\sigma$\verb|)|.
        \STATE $\hat{\mathcal{D}}^n, \hat{\mathcal{D}}^f \leftarrow$ \verb|SampleSelection(|$\mathcal{D}$, $H^{n}$, $H^{f}$, $S$, $C$\verb|)|.
        \STATE $\{w_k\}_{k=1}^K \leftarrow$ \verb|PLMScoring(|$H^{n}$, $\{\mathcal{D}_k\}_{k=1}^K$\verb|)|.
    \ENDFOR
    \STATE $m \leftarrow$ \verb|STMTraining(|$\mathcal{D}$, $m_{(0)}$, $\eta$\verb|)|.
\end{algorithmic}
\end{algorithm}

\subsection{Differentially Private Top-$Q$ Voting} \label{subsec:method_step2} 
As shown in \cref{subfig:pe_random_fid_and_acc}, with limited real private samples, noisy estimations of the real private data distribution cause the original PE algorithm to fail in generating synthetic samples that resemble private data. Our aim is to improve distribution estimations and generation guidance in this scenario.
To achieve this, unlike previous works~\cite{lin2024differentially,xie2024differentially,hou2024pretext} that assign only $1$ vote per private sample, we propose a Top-$Q$ voting mechanism with decaying weights. This approach maximizes the use of limited private samples by giving weighted votes to the Top-$Q$ nearest and furthest synthetic samples relative to the private sample.
Specially, we first compute the pair-wise distance between each of the private samples $(\mathbf{z}_j,u_j)\in \mathcal{B}$ and each synthetic sample $(\mathbf{x}_{i},y_{i}) \in \mathcal{D}$ if they possess the same label, i.e. $y_i=u_j$. 
Using $\ell_2$ distance as measurement, we have:
\begin{equation} \label{eq:distance}
\begin{split}
    d(\mathbf{z}_{j},\mathbf{x}_{i})=||\varphi(\mathbf{z}_{j})-\varphi(\mathbf{x}_{i})||_2 \,, \\
    \forall \,\, j=1,\dots,M; \,\, (\mathbf{x}_{i},y_i) \in \mathcal{D}^{[u_j]} \, ,
\end{split}
\end{equation}
where $\varphi$ denotes a pre-trained sentence embedding model and $\mathcal{D}^{[u_j]}$ denotes the subset of $\mathcal{D}$ which has a label that equals to $u_j$. Next, we use each private sample $(\mathbf{z}_j,u_j)\in \mathcal{B}$ to vote for its Top-$Q$ nearest and Top-$Q$ furthest synthetic samples within $\mathcal{D}^{[u_j]}$ based on \cref{eq:distance}.
The indices of the synthetic samples selected by each $(\textbf{z}_{j},u_j) \in \mathcal{B}$ are:
\begin{equation} \label{eq:distance_based_scoring}   
\begin{split}
    [n_{j,1},\dots,n_{j,Q}] &\leftarrow \arg\mathrm{top}Q\mathrm{Smallest} \left( \, d(\textbf{z}_j, \textbf{x}_{i})_{(\mathbf{x}_{i},y_i)\in \mathcal{D}^{[u_j]}} \right), \\
    [f_{j,1},\dots,f_{j,Q}] &\leftarrow \arg\mathrm{top}Q\mathrm{Largest} \left( \, d(\textbf{z}_j, \textbf{x}_{i})_{(\mathbf{x}_{i},y_i)\in \mathcal{D}^{[u_j]}} \right) .
\end{split}
\end{equation}
where functions $\arg\mathrm{top}Q\mathrm{Smallest}$ and $\arg\mathrm{top}Q\mathrm{Largest}$ return the indices of the Top-$Q$ samples with the smallest and largest $d(\textbf{z}_j, \textbf{x}_{i})$, respectively, with $n_{j,1},\dots,n_{j,Q},f_{j,1},\dots,f_{j,Q}$ denoting the index of 
 selected samples. 
To utilize the relative ranking information, as well as to guarantee a controllable function sensitivity for DP protection, we assign decreasing voting weights $1,\frac{1}{2},\dots,\frac{1}{2^{Q-1}}$ to each of the Top-$Q$ selected samples when producing the voting histograms, \textit{Nearest Histogram} $H^{n}$ and \textit{Furthest Histogram} $H^{f}$. This can be formulated as:
\begin{equation} \label{eq:voting_q}
\begin{split}
    H^{n}[n_{j,q}] \leftarrow H^{n}[n_{j,q}]+\frac{1}{2^{q-1}}, \, H^{f}[f_{j,q}] \leftarrow H^{f}&[f_{j,q}]+\frac{1}{2^{q-1}} \\
    \forall (\textbf{z}_j,u_j)\in\mathcal{B}, \, \forall \, q \in [1,\dots,Q],
\end{split}
\end{equation}
with $H^n,H^f$ each initialized as $[0,\dots,0]$ of length $|\mathcal{D}|$. 

To further guarantee $(\epsilon,\delta)$-DP for private samples,  Gaussian noises following $\mathcal{N}(0,\sigma^2)$ with $\sigma=4\frac{\sqrt{2\log{\left(1.25/\delta\right)}}\sqrt{T-1}}{\epsilon}$ are added to $H^{n},H^{f}$:
\begin{equation}\label{qq:dp_noise_addition}
    H^{n} \leftarrow H^{n}+\mathcal{N}(0,\sigma^2I_{|\mathcal{D}|}), \, H^{f} \leftarrow H^{f}+\mathcal{N}(0,\sigma^2I_{|\mathcal{D}|})\,,
\end{equation}
where $I_{|\mathcal{D}|}$ represents the identity matrix of size $|\mathcal{D}|\times|\mathcal{D}|$.

Based on $H^{n},H^{f}$, for each category $c$,  we select low-quality samples with the highest votes in $H^f$ (largest distance to private samples in $\mathcal{B}$), denoted as $\hat{\mathcal{D}}^{f,[c]}$, 
alongside high-quality samples with the highest votes in $H^n$ (nearest to private samples in $\mathcal{B}$), denoted as $\hat{\mathcal{D}}^{n,[c]}$, 
following:
\begin{equation} \label{eq:incontext_sample_selection}
\begin{split}
    H^{n,[c]} =& \left\{ H^n[i] \,\big|\, (\mathbf{x}_{i},y_i) \in \mathcal{D}^{[c]} \right\}, \\
    H^{f,[c]} =& \left\{ H^f[i] \,\big|\, (\mathbf{x}_{i},y_i) \in \mathcal{D}^{[c]} \right\}, \\
    \hat{\mathcal{D}}^{n,[c]} =& \left\{ (\mathbf{x}_i,y_i) \in \mathcal{D}^{[c]} \,\big|\, H^n[i] \text{ is in the top-}S \text{ values of } H^{n,[c]} \right\}, \\
    \hat{\mathcal{D}}^{f,[c]} =& \left\{ (\mathbf{x}_i,y_i) \in \mathcal{D}^{[c]} \,\big|\, H^f[i] \text{ is in the top-}S \text{ values of } H^{f,[c]} \right\}, 
\end{split}
\end{equation}
where $S$ is the amount of samples to select and $H^{n,[c]},H^{f,[c]}$ denote the sets of the nearest and furthest voting results of samples belonging to category $c$. 
$\hat{\mathcal{D}}^{n} = \bigcup_{c=1}^C \hat{\mathcal{D}}^{n,[c]}, \hat{\mathcal{D}}^{f} = \bigcup_{c=1}^C \hat{\mathcal{D}}^{f,[c]}$ are the total sets of high- and low-quality samples respectively.
Note that we do not limit the origin of the selected samples, and synthetic samples generated by different PLMs can all be included in $\hat{\mathcal{D}}^n$ and $\hat{\mathcal{D}}^f$.

\subsection{PLM Importance Weighting} 
\label{subsec:method_step3}

Previous studies on API-based multi-PLM fusion~\cite{li2024more,zou2024fusegen} 
often treat involved PLMs equally.
However, as shown in \cref{subfig:pe_variance} and \cref{fig:pe_q8_contrast_comparison} in \cref{subsec:appendix_single_plm_topQ_contrast}, different PLMs exhibit varying generation capabilities, leading to uneven synthetic data quality. This encourages 
assigning customized weights for each PLM to enhance their contributions. Therefore, we introduce a PLM weighting strategy based on the quality of their generated synthetic data, which is measured by their similarity to private samples.

Since the \textit{Nearest Histogram} $H^{n}$ obtained in \cref{eq:voting_q} quantifies the similarity between each synthetic sample and private samples, we simply aggregate the histogram values of each synthetic sample with source PLM $\mathcal{P}_k$ to obtain the weight $w_k$ of the PLM $\mathcal{P}_k$ for the upcoming generation iteration. That is,
\begin{equation} \label{eq:weight_calculation}
\begin{split}  
    s_{i} &= \frac{H^n[i] }{\sum_{i'=1}^{|\mathcal{D}|} H^n[i']},\\
    w_k &= \frac{\sum_{(\textbf{x}_{i},y_{i})\in\mathcal{D}_k} s_{i}}{\frac{|\mathcal{D}_{k}|}{\sum_{k'=1}^K |\mathcal{D}_{k'}|}} = \frac{\sum_{(\textbf{x}_{i},y_{i})\in\mathcal{D}_k} s_{i}}{|\mathcal{D}_{k}| / |\mathcal{D}|} \, .
\end{split}
\end{equation}

\subsection{Cross-PLM Contrastive In-context Learning (ICL)} \label{subsec:method_step4}
Inspired by the observation that low-quality samples still exist in DP synthetic dataset given by PE (see \cref{tab:examples_bad_samples} in \cref{subsec:appendix_good_bad_samples}), we select cross-PLM contrastive samples from $\hat{\mathcal{D}}^n$ and $\hat{\mathcal{D}}^f$ (obtained in \cref{subsec:method_step3}), and use them to create a \textit{contrastive task-related label-descriptive prompt} $\mathcal{T}(\cdot)$ to perform cross-PLM contrastive ICL. 
$\mathcal{T}(\cdot)$ describes the task, provides category description, and contains explicit contrastive instructions for high- and low-quality samples. It contains the following sequential instructions: $(1)$ analyze the difference between low- and high-quality samples; $(2)$ ensure the new sample is better in quality and closer to real private distribution than the high-quality samples, and is further away from the low-quality samples than the high-quality samples; $(3)$ generate a new sample which is diverse in expression compared to the given high-quality samples. 
Note that to improve the generation diversity, for each generation we perform random sample selection to draw $50\%$ of samples respectively from $\hat{\mathcal{D}}^{f,[c]}$ and $\hat{\mathcal{D}}^{n,[c]}$ to construct the final in-context samples for $\mathcal{T}(c)$. 
Also, different from PE algorithms series, we choose not to 
vary 
one existed synthetic sample each time, but to encourage diverse sample generation using $S$ demonstrations at once. 
Prompt examples can be found in \cref{tab:appendix_prompt} in \cref{sec:appendix_prompt}.

\subsection{WASP in Federated Data Setting} \label{subsec:multi_pdp}

So far we have built our algorithms under single data-party setting, which can be easily extended to federated data scenario~\cite{hou2024pretext}, where each data party possesses an insufficient amount of private data and collaborates on private tasks. This scenario is very common in the real world, such as collaborations between medical companies. 
In this setting, we consider $L$ data parties $\{\mathcal{C}_l\}_{l=1}^L$, each possessing a real private dataset $\mathcal{B}_l = \left\{(\mathbf{z}_{l,j},y_{l,j})\right\}_{j=1}^{M_l}$ of size $M_l$. These data parties aim to collaboratively generate a DP synthetic dataset  while preserving local data privacy.
The full algorithm is provided in \cref{alg:algorithm_full_functions}.

When extending to federated data setting, each party $\mathcal{C}_l$ uses its local private samples in $\mathcal{B}_l$ to perform DP Top-Q voting with $\sigma=4\frac{\sqrt{2\log{\left(1.25/\delta\right)}}\sqrt{T-1}}{\epsilon\sqrt{L}}$ to guarantee privacy. The produced local nearest and furthest voting histograms $\{H^{n}_l\}_{l=1}^L,\{H^{f}_l\}_{l=1}^L$ are then securely aggregated~\cite{bonawitz2016practical} before sent to a central server following: 
\begin{equation}\label{eq:sum_histogram}
\small
    H^{n}\leftarrow\sum_{l=1}^{L} H^{n}_l, \, H^{f}\leftarrow\sum_{l=1}^{L} H^{f}_l.
\end{equation}
We adopt an honest-but-curious threat model where the server only has access to the aggregated histograms $H^{n}$ and $H^{f}$, but not individual ones. We also assume that all data parties participate in the aggregation and therefore aims to ensure sample-level $(\epsilon,\delta)$-DP of $\mathcal{B}$ (see \cref{sec:appendix_privacy_analysis}). Note that, WASP can be easily extended 
to ensure user-level DP, with discussions and results included in \cref{subsec:appendix_user_level_dp_results}.


\section{Experiments} \label{sec:experiments}

\subsection{Settings} \label{subsec:experimental_settings}
\textbf{Models.} In this work, 6 open-source PLMs and 3 closed-source PLMs are considered. Open-source PLMs include GPT-2-xl (GPT-2)~\cite{radford2019language}, Llama-2-7b-chat-hf (Llama-2)~\cite{touvron2023llama2}, Vicuna-7b-1.5v (Vicuna)~\cite{vicuna2023}, OPT-6.7b (OPT)~\cite{zhang2022opt}, ChatGLM3-6b-base (ChatGLM3)~\cite{du2022glm}, and Flan-T5-xl (Flan-T5)~\cite{chung2022scaling}. Close-source PLMs include GPT-3.5-turbo-instruct (GPT-3.5)~\cite{openai2021gpt3-5}, GPT-4-turbo-preview (GPT-4)~\cite{openai2023gpt4}, and GPT-4o~\cite{hurst2024gpt4o}. For STM, we use pre-trained bert-base-uncased (BERT) model and further fine-tune it on downstream classification tasks using $\mathcal{D}$. 
We use sentence-t5-base~\cite{ni2022sentencet5base} as the embedding model $\varphi$.

\textbf{Datasets.} We evaluate on $6$ widely used tasks: $1)$ IMDb~\cite{maas2011learning_imdb} ($2$ categories) for movie-review semantic analysis task; $2)$ Yelp-Category~\cite{yelpopendataset} ($10$ categories) for business-review item field classification task; $3)$ Yelp-Rating~\cite{yelpopendataset} ($5$ categories) for business-review rating classification task; $4)$ Openreview-Category~\cite{xie2024differentially} ($12$ categories) for paper-review classification by research area task; $5)$ Openreview-Rating~\cite{xie2024differentially} ($5$ categories) for paper-review classification by review rating task; and $6)$ Banking ($10$ categories selected from Banking77~\cite{Casanueva2020banking77}) for online-banking queries field classification task. $\mathcal{B}$ is randomly drawn from the training sets of these datasets with their test sets used to evaluate trained STM. 

\textbf{Baselines.} We compare the WASP framework to $4$ baselines: $1)$ Aug-PE~\cite{xie2024differentially}, the original PE algorithm specialized for text modality; $2)$ Pre-Text~\cite{hou2024pretext}, which applies PE to federated private data setting; $3)$ OnlyPrivate, the centralized training method relying merely on $\mathcal{B}$ without DP ($\epsilon=\infty$), which provides a performance upper-bound of using no synthetic data; $4)$ FuseGen~\cite{zou2024fusegen}, which generates synthetic data in a zero-shot manner without accessing private samples. 

\begin{table*}[t]
\caption{Evaluation of downstream STM accuracy using $6$PLMs, $L=1$. 
\textbf{Best} and \underline{second best} results are marked.
}
\label{tab:main_results_1}
\begin{center}
\begin{small}
    \begin{tabular}{c|c|c|cc||c|cc|cc|c}
    \toprule
        \multicolumn{2}{c|}{} & \multirow{2}{*}{Privacy} & \multirow{2}{*}{\,\,$|\mathcal{B}|$\,\,} & \multirow{2}{*}{$|\mathcal{D}|$} & \multirow{2}{*}{IMDb} & \multicolumn{2}{c|}{Yelp} & \multicolumn{2}{c|}{Openreview} & \multirow{2}{*}{Banking} \\
        \multicolumn{2}{c|}{} & ~ & ~ & ~ & ~ & Category & Rating & \,\,Area\,\, & Rating & ~ \\
    \midrule
        \multicolumn{2}{c|}{OnlyPrivate} & $\epsilon=\infty$ & $100$ & - & 50.00 & 5.69 & 35.57 & 6.56 & 22.20 & 13.75 \\
    \midrule
        \multicolumn{2}{c|}{FuseGen} & Absolutely Private & - & $6,000$ & \underline{89.07} & \underline{63.38} & 57.96 & 24.70 & 34.57 & 78.75 \\
    \midrule
        \multirow{6}{*}{\shortstack{Aug-PE}} & GPT-2 & ($4.0,1\times10^{-5}$)-DP & $100$ & $6,000$ & 85.38 & 62.33 & 45.28 & 31.45 & 24.12 & 75.63 \\
         & Llama-2 & ($4.0,1\times10^{-5}$)-DP & $100$ & $6,000$ & 85.77 & 60.18 & 47.42 & 32.67 & 34.78 & 84.63 \\
         & Vicuna & ($4.0,1\times10^{-5}$)-DP & $100$ & $6,000$ & 82.76 & 63.28 & 54.42 & 32.27 & 30.66 & 86.75 \\
         & OPT & ($4.0,1\times10^{-5}$)-DP & $100$ & $6,000$ & 83.86 & 62.71 & 50.81 & \underline{34.64} & 25.30 & 79.25 \\
         & ChatGLM3 & ($4.0,1\times10^{-5}$)-DP & $100$ & $6,000$ & 85.82 & 55.06 & 55.17 & 33.81 & 32.49 & \underline{88.50} \\
         & Flan-T5 & ($4.0,1\times10^{-5}$)-DP & $100$ & $6,000$ & 89.00 & 62.06 & \underline{58.69} & 34.54 & \underline{35.42} & 81.25 \\
    \midrule
        \multicolumn{2}{c|}{WASP (Ours)} & ($4.0,1\times10^{-5}$)-DP & $100$ & $6,000$ & \textbf{89.52} & \textbf{63.91} & \textbf{61.21} & \textbf{34.99} & \textbf{37.10} & \textbf{88.75} \\
    \bottomrule
    \end{tabular}
\end{small}
\end{center}
\end{table*}

\begin{table*}[t]
\caption{Evaluation of downstream STM accuracy using $6$ PLMs, $L=10$. 
\textbf{Best} and \underline{second best} results are marked.
}
\label{tab:main_results_10}
\begin{center}
\begin{small}
    \begin{tabular}{c|c|c|cc||c|cc|cc|c}
    \toprule
        \multicolumn{2}{c|}{} & \multirow{2}{*}{Privacy} & \multirow{2}{*}{\,\,$|\mathcal{B}|$\,\,} & \multirow{2}{*}{$|\mathcal{D}|$} & \multirow{2}{*}{IMDb} & \multicolumn{2}{c|}{Yelp} & \multicolumn{2}{c|}{Openreview} & \multirow{2}{*}{Banking} \\
        \multicolumn{2}{c|}{} & ~ & ~ & ~ & ~ & Category & Rating & \,\,Area\,\, & Rating & ~ \\
    \midrule
        \multicolumn{2}{c|}{OnlyPrivate} & $\epsilon=\infty$ & $100$ & - & 50.00 & 5.90 & 38.76 & 8.86 & 23.55 & 16.75 \\
    \midrule
        \multicolumn{2}{c|}{FuseGen} & Absolutely Private & - & $6,000$ & \underline{89.07} & 63.38 & 57.96 & 24.70 & 34.57 & 78.75 \\
    \midrule
        \multirow{6}{*}{\shortstack{Pre-Text}} & GPT-2 & ($4.0,1\times10^{-5}$)-DP & $100$ & $6,000$ & 85.87 & 62.58 & 46.25 & 37.13 & 24.45 & 76.25 \\
         & Llama-2 & ($4.0,1\times10^{-5}$)-DP & $100$ & $6,000$ & 86.09 & 60.20 & 51.11 & 34.24 & 36.24 & 85.38 \\
         & Vicuna & ($4.0,1\times10^{-5}$)-DP & $100$ & $6,000$ & 83.52 & \underline{64.11} & 54.76 & 36.38 & 30.88 & 86.13 \\
         & OPT & ($4.0,1\times10^{-5}$)-DP & $100$ & $6,000$ & 83.98 & 63.65 & 52.44 & 37.67 & 24.73 & 79.75 \\
         & ChatGLM3 & ($4.0,1\times10^{-5}$)-DP & $100$ & $6,000$ & 86.32 & 60.24 & 56.94 & 38.14 & 33.35 & \underline{89.38} \\
         & Flan-T5 & ($4.0,1\times10^{-5}$)-DP & $100$ & $6,000$ & 89.02 & 62.82 & \underline{61.04} & \underline{38.31} & \underline{36.53} & 81.75 \\
    \midrule
        \multicolumn{2}{c|}{WASP (Ours)} & ($4.0,1\times10^{-5}$)-DP & $100$ & $6,000$ & \textbf{89.65} & \textbf{64.34} & \textbf{61.46} & \textbf{40.47} & \textbf{37.60} & \textbf{89.63} \\
    \bottomrule
    \end{tabular}
\end{small}
\end{center}
\end{table*}

\textbf{Implementation Details.} 
By default, we use $100$ private samples ($M=100$) for main experiments. For federated data ($L>1$) scenario, we use $L=10$ private data parties which control $300$ private samples ($M=\sum_{l=1}^{10} |\mathcal{B}_l|=300$) altogether. To better align with real-world scenarios, each participating data-party controls private datasets that are \textit{non-i.i.d.} to each other, and aggregate to an unbalanced dataset. 
We follow Dirichlet Partition~\cite{yurochkin2019bayesian,hsu2019measuring,zhang2023fedala} to distribute private samples to each party
with parameter $\alpha=1.0$.
For the DP synthetic dataset, we generate a total of $6,000$ samples from all participating PLMs within $5$ iteration. 
The notion of DP is sample-level DP unless otherwise stated.

\subsection{Main Results} \label{subsec:main_results}

\textbf{Single Data Party Setting.} 
Experimental results using $K=6$ open-source PLMs and $3$ closed-source PLMs are provided in \cref{tab:main_results_1,tab:close_plm_main_results}, which show that WASP outperforms all baseline methods across different tasks, demonstrating its superiority. As expected, the closed-source GPT series (see \cref{tab:close_plm_main_results}), being powerful models, outperform their open-source counterparts (see \cref{tab:main_results_1}) when using baseline method Aug-PE. 

For all tasks, with limited private samples, Aug-PE performs poorly when using improper single PLM, e.g. using OPT for IMDb and using GPT-2 for Openreview-Rating.
Differently, WASP performs consistently well across tasks, and achieves a lower FID value compared to baselines (see \cref{fig:appendix_fid_comparison} in \cref{subsec:appendix_fid_comparison}), verifying its effectiveness under limited private sample setting.
Also, the best performing PLM model varies across tasks for Aug-PE, 
highlighting the arbitrary nature of PLM selection. In contrast, WASP consistently achieves better performance across tasks, making it PLM-agnostic without requiring prior-knowledge for selecting specific PLMs for collaboration.

On the other hand, comparing with FuseGen, a baseline under zero-shot setting where private samples are inaccessible, WASP leverages real private samples and utilizes a more targeted PLM importance weighting method,  therefore achieving better performance.
Moreover, the notably poor performance of ``OnlyPrivate'' shows that the trained STM  relying merely on  private dataset $\mathcal{B}$ is nearly unusable, even without applying DP during training which can further degrade STM performance.

\begin{table}[]
\centering
\caption{Evaluation of downstream STM accuracy using $3$ closed-source PLMs, $L=1$ with the same DP setting in \cref{tab:main_results_1}.
\textbf{Best} and \underline{second best} results are marked.
}
\label{tab:close_plm_main_results}
\begin{small}
    \resizebox{1\linewidth}{!}{
    \begin{tabular}{c||c|c|ccc|c}
    \toprule
        \multirow{2}{*}{~} & \multirow{2}{*}{\shortstack{Only\\Private}} & \multirow{2}{*}{\shortstack{FuseGen}} & \multicolumn{3}{c|}{Aug-PE} & \multirow{2}{*}{\shortstack{WASP\\(Ours)}} \\
    \cline{4-6}
    \\[-0.8em]
        ~ & ~ & ~ & GPT-3.5 & GPT-4 & GPT-4o & ~ \\
    \midrule
        \multirow{2}{*}{\shortstack{Yelp-\\Rating}} & \multirow{2}{*}{35.57} & \multirow{2}{*}{61.36} & \multirow{2}{*}{60.90} & \multirow{2}{*}{61.02} & \multirow{2}{*}{\underline{62.06}} & \multirow{2}{*}{\textbf{64.48}} \\
        ~ & ~ & ~ & ~ & ~ & ~ & ~ \\
    \bottomrule
    \end{tabular}
    }
\end{small}
\end{table}

\textbf{Federated Data Setting.}
We also conduct experiments under distributed federated data setting, with $L=10$ and $M=300$ total number of private samples. Results in \cref{tab:main_results_10} show that WASP consistently achieves better performance across different tasks and settings compared to Pre-Text, a baseline designed for federated data. This further demonstrates the effectiveness of WASP when extended to federated data setting.
Additional results on communication cost comparison is given in \cref{tab:upload_download_comparison} in \cref{subsec:appendix_communication_overhead}, where we show that the communication increase caused by uploading additional histograms by our method is minimal. 

\subsection{Ablation Studies}
\begin{figure}[!tb]
\vspace{-1em}
    \centering
     \subfigure[Effect of $K$]{
         \begin{minipage}[t]{0.47\linewidth}
         \centering
         \includegraphics[width=1\linewidth]{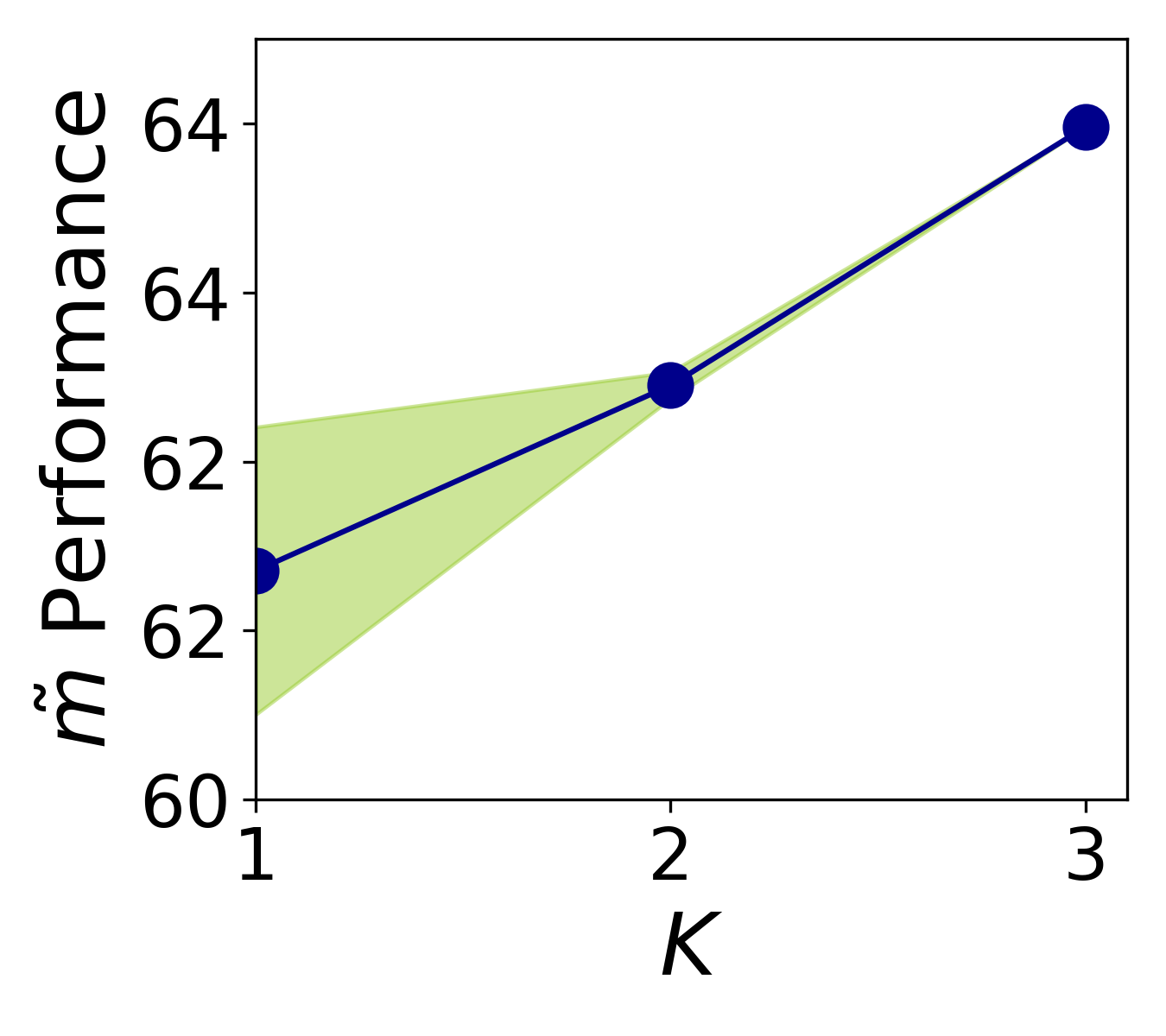}
         \vspace{-0.5em}
         \label{fig:errorbar_increase_k}
         \end{minipage}
     }
    \subfigure[Pair-wise Comparison]{
         \begin{minipage}[t]{0.43\linewidth}
         \centering
         \includegraphics[width=1\linewidth]{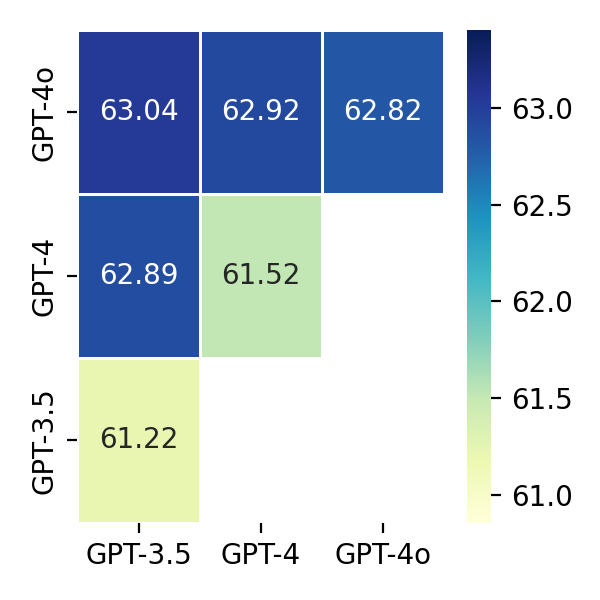}
         \vspace{-0.5em}
         \label{fig:pair_wise}
         \end{minipage}
     }
     \vspace{-0.3em}
     \caption{Evaluation of downstream STM accuracy using Yelp-Rating dataset with $K=1,2,3$ closed-source PLMs, $L=1$ under $(4.0,1\times10^{-5})$-DP setting. In (b), results on the diagnose are with $K=1$ and others are with $K=2$.}
    \label{fig:increase_of_k_closed_source_plms}
\vspace{-1em}    
\end{figure}

\textbf{\# PLMs ($K$).} \label{s titleubsec:experiments_K}
We first study the impact of the number of PLMs ($K$) on the final STM performance. Results of using $1,2,3$ closed-source PLMs under $L=1$ and $(4.0,1\times10^{-5})$-DP settings are reported in \cref{fig:errorbar_increase_k}. We can see that the performance of $m$ increases simultaneously with the increase of $K$ while the randomness (STD) decreases.
This indicates that the randomness in the performance of the synthetic dataset can be mitigated by incorporating more PLMs into WASP, which simultaneously increases the performance expectations. 

We also display the pair-wise combination ($K=2$) results of the $3$ closed-source PLMs under $L=1$ and $(4.0,1\times10^{-5})$-DP settings in \cref{fig:pair_wise}. In this figure, any pair-wise collaboration ($K=2$) outperforms either participating single-PLM alone (diagnose in \cref{fig:pair_wise}), demonstrating that WASP performs better using the whole set of available PLMs than using only a subset of them. These findings show that WASP’s improvements are PLM-agnostic, independent of any single PLM's inherent task capabilities. Consequently, WASP effectively mitigates the risk of selecting the optimal PLM by harnessing the collective strengths of all participating models. 


\textbf{Contrastive ICL \& PLM Importance Weighting. 
} \label{subsec:experiments_ablatio_weighting_and_contrast}
To evaluate the effectiveness of our proposed \textit{Contrastive In-context Learning} and \textit{PLM Importance Weighting} methods, we conduct ablation experiments to see how these components impact the final STM performance. Results are reported in \cref{tab:plm_weighting_comparison}. 
By removing \textit{Contrastive In-context Learning} (labeled as ``w/o PLM Contrastive Prompting'' in \cref{tab:plm_weighting_comparison}), we only select high-quality samples for the prompt (details in \cref{tab:appendix_prompt}). This leads to a $0.31\%$ decrease in STM performance on the easier IMDb task, and a much larger $1.56\%$ and $0.92\%$ decrease on the more challenging Yelp-Rating and Openreview-Rating tasks. This highlights the importance of using low-quality samples as feedback demonstrations to encourage the PLMs avoid generating low-quality DP synthetic samples.

On the other hand, by removing \textit{PLM Importance Weighting} (labeled as ``w/o PLM Importance Weighting'' in \cref{tab:plm_weighting_comparison}), $w_k=1/K$ within each generation iteration, indicating that each  PLM generates equal amount of samples across iterations. Similarly, results indicate a $0.35\%$ decrease in STM performance on the easier IMDb task and a $2.27\%$ and $1.57\%$ decline on the more challenging Yelp-Rating and Openreview-Rating tasks. This underlines the effectiveness of weighted aggregation of PLMs with varying degrees of reliance on their capabilities for specific task. 
Furthermore, these results demonstrate that by generating better DP synthetic data, WASP is more effective than baselines when faced with more challenging tasks. 

\begin{table}[]
    \centering
    \caption{Comparison of downstream STM accuracy under w/ and w/o Contrastive In-context Learning and Private Data Assisted PLM Importance Weighting setting using $6$ open-source PLMs, $L=1$ with $(4.0, 1\times10^{-5})$-DP.}
    \label{tab:plm_weighting_comparison}
\begin{small}
    \begin{tabular}{c||cc|c}
    \toprule
        ~  & {\shortstack{w/o PLM\\Contrastive\\Prompting}} & {\shortstack{w/o PLM\\Importance\\Weighting}} & {\shortstack{WASP\\(Ours)}} \\
    \midrule
        IMDb & 89.21 & 89.17 & 89.52\\
        Yelp-Rating & 59.65 & 58.94 & 61.21 \\
        Openreview-Rating & 36.18 & 35.53 & 37.10 \\
    \bottomrule
    \end{tabular}
\end{small}
\end{table}

\textbf{\# Votes ($Q$) by Each Private Sample.} \label{subsec:experiments_Q}
To better estimate private sample distribution  with limited private samples, WASP exploits each private sample by increasing the amount of votes each private sample gives out (from $Q=1$ in previous works to $Q=8$). Here we investigate how the change in $Q$ impacts  STM performance. Results in \cref{tab:voting_count_q} indicate that STM performance improves with higher values of $Q$, but the improvement diminishes at larger $Q$ ($Q>8$). 
This underscores the strength of our idea in increasing the utility of each private sample to achieve a more accurate private sample distribution estimation, particularly in scenarios with limited available private samples.

\begin{table}[]
\centering
\caption{Evaluation of downstream STM accuracy using $6$ open-source PLMs, $L=1$ with $(4.0, 1\times10^{-5})$-DP under different $Q$.}
\label{tab:voting_count_q}
\begin{small}
    \resizebox{1\linewidth}{!}{
    \begin{tabular}{c||ccccc}
    \toprule
         & $Q=1$ & $Q=2$ & $Q=4$ & $Q=8$ & $Q=16$ \\
    \midrule
        IMDb & 89.02 & 89.15 & 89.39 & 89.52 & 89.60 \\
        Yelp-Rating & 58.74 & 58.92 & 59.24 & 61.21 & 61.42 \\
    \bottomrule
    \end{tabular}
    }
\end{small}
\end{table}

\begin{figure}
    \centering
    \includegraphics[width=0.99\linewidth]{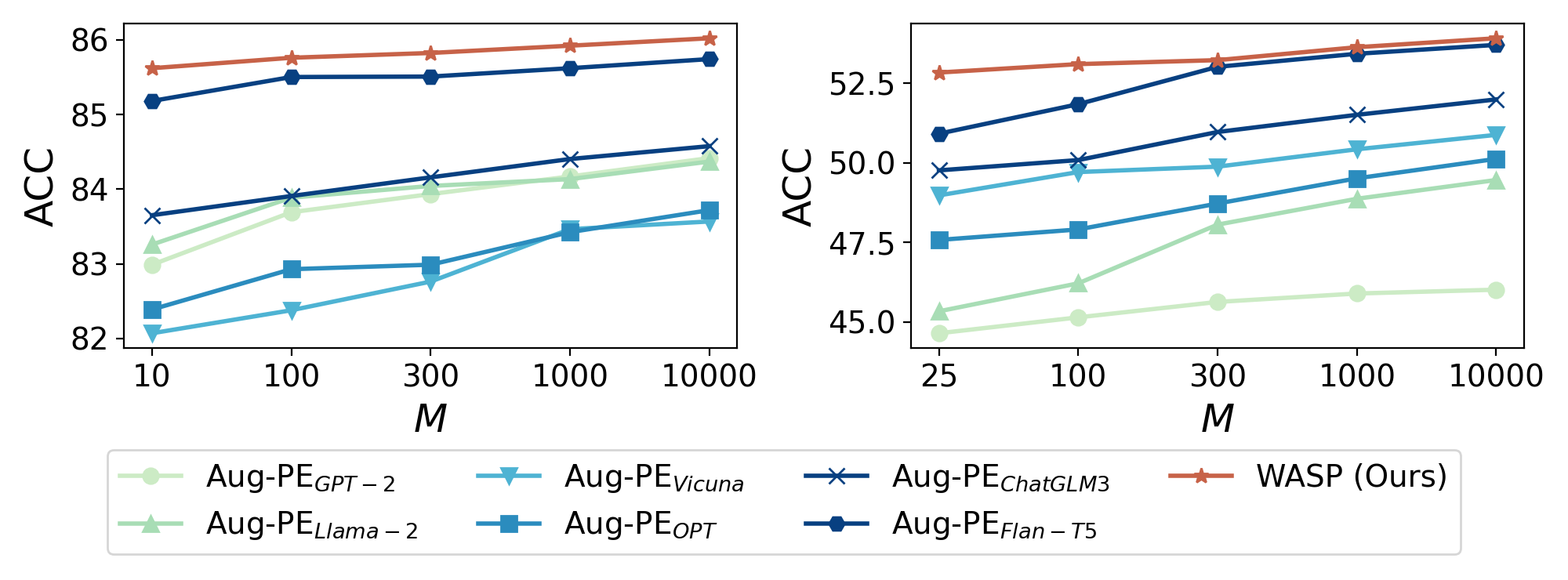}
    \vspace{-0.5em}
    \caption{Comparison of downstream STM accuracy using different number of private samples ($M$) from the training set of IMDb and Yelp-Rating datasets using $6$ open-source PLMs, $L=1$ with $(4.0,1\times10^{-5})$-DP.}
    \label{fig:sensitive_m}
\end{figure}

\textbf{Sensitive Analysis of \# Private Samples ($M$).} \label{subsec:experiments_M}
We also investigate the impact of $M$ on WASP. 
In \cref{fig:sensitive_m}, we compare the baseline Aug-PE method with WASP across different values of $M$. Results show that WASP consistently outperforms Aug-PE across different values of $M$ for all PLMs, and the performance gaps at smaller $M$ values ($M<1000$) are much greater, 
underscoring the effectiveness of WASP in limited private data scenarios. 

\textbf{Different Private Budget ($\epsilon$).} \label{subsec:experiments_epsilon}
As illustrated in \cref{tab:dp_alpha}, STM performance using WASP gradually declines from $89.96\%$ to $89.36\%$ for IMDb and from $62.02\%$ to $60.94\%$ for Yelp-Rating as the privacy budget $\epsilon$ decreases from $\infty,8.0,4.0$ to $1.0$, similar to that of Aug-PE when using the best performing single PLM for each task. This indicates that WASP scales well with $\epsilon$ and maintains high performance even under tight privacy constraints, just like baseline method. 

\begin{table}[]
\centering
\caption{Evaluation of downstream STM accuracy using $6$ open-source PLMs, $L=1$ under different DP budget setting with $\delta=1\times10^{-5}$. The best performing PLM is used for Aug-PE evaluation, i.e. Flan-T5 for both tasks.}
\label{tab:dp_alpha}
\begin{small}
    \resizebox{1\linewidth}{!}{
    \begin{tabular}{c|c||cccc}
    \toprule
         ~ & ~ & $\epsilon=\infty$ & $\epsilon=8.0$ & $\epsilon=4.0$ & $\epsilon=1.0$ \\
    \midrule
        \multirow{2}{*}{IMDb} & WASP & 89.96 & 89.77 & 89.52 & 89.36 \\
        ~ & Aug-PE & 89.48 & 89.23 & 89.00 & 88.72 \\
        \midrule
        \multirow{2}{*}{\shortstack{Yelp-\\Rating}} & WASP & 62.02 & 61.54 & 61.21 & 60.94 \\
        ~ & Aug-PE & 59.62 & 59.12 & 58.69 & 58.59 \\
    \bottomrule
    \end{tabular}
    }
\end{small}
\end{table}


\section{Conclusion and Future Work}
In this work, we introduce a novel DP synthetic data generation framework, WASP, which leverages the collaborative capabilities of multiple PLMs to address real-world scenarios with limited private samples, 
 while observing differential privacy. 
Extensive experiments  across $6$ tasks demonstrate that WASP is highly effective, PLM-agnostic, scalable with respect to privacy budgets, and superior in challenging scenarios, making it a practical and scalable solution for real-world applications.

Possible future work points to more precise sample-level weighting or selection to further improve the quality of the DP synthetic dataset, as well as verifying the effectiveness of WASP on non-classification tasks.


\bibliographystyle{ACM-Reference-Format}
\bibliography{ref}

\appendix
\newpage

\section{Contrastive Prompts and Non-contrastive Prompts} \label{sec:appendix_prompt}
\begin{table*}[!htb]
\centering
\caption{Prompt used for synthetic dataset generation. Due to clarity, we omit the words in the parentheses in the labels of Openreview-Category and the attributes of Openreview-Rating.}
\label{tab:appendix_prompt}
\vspace{0.3em}
    \resizebox{0.998\linewidth}{!}{
    \begin{tabular}{c|c|p{11.5cm}|c|c}
    \toprule
        Dataset (task) & prompt type & prompt & label & attribute \\
    \midrule
        \multirow{2}{*}{\shortstack{IMDb\\(semantic analysis\\of movie review)}} & w/o Contrastive & {``The movie review is: \textit{<sample\_1>}\emph{\textbackslash n}The movie review is: \textit{<sample\_2>}\emph{\textbackslash n}...\emph{\textbackslash n}The movie review is: \textit{<sample\_S>}\emph{\textbackslash n}\emph{\textbackslash n}Based on the above movie reviews, a new movie review also in \textit{\textbf{<label>}} sentiment but diverse in the expression compared to the above given samples is: ''} & \multirow{2}{*}{\textit{\textbf{positive / negative}}} & \multirow{2}{*}{None} \\
        \cline{2-3}
        \\[-1em]
        ~ & w/ Contrastive & ``A bad movie review is: \textit{<sample\_1>}\emph{\textbackslash n}...\emph{\textbackslash n}A bad movie review is: \textit{<sample\_$\lfloor$S/2$\rfloor$>}\emph{\textbackslash n}A good movie review is: \textit{<sample\_$\lfloor$S/2$\rfloor$+1>}\emph{\textbackslash n}...\emph{\textbackslash n}A good movie review is: \textit{<sample\_S}\emph{\textbackslash n}\emph{\textbackslash n}Based on the above examples of bad and good movie reviews in \textit{\textbf{<label>}} sentiment, analyze the differences between the bad and good reviews. Generate a new positive movie review that is diverse in expression compared to the given good reviews. Ensure that the new review is further refined than the good reviews while maintaining the \textit{\textbf{<label>}} sentiment and clarity, making the good reviews appear to lie midway between the new review and the bad reviews. The new \textit{\textbf{<label>}} movie review is: '' & ~ & ~ \\
        \hline
        \\[-1em]
        \multirow{2}{*}{\shortstack{Yelp-Category\\(field classification\\of business review)}} & w/o Contrastive & The business review is: \textit{<sample\_1>}\emph{\textbackslash n}The business review is: \textit{<sample\_2>}\emph{\textbackslash n}...\emph{\textbackslash n}The business review is: \textit{<sample\_S>}\emph{\textbackslash n}\emph{\textbackslash n}Based on the above business reviews belonging to the category of \textit{\textbf{<label>}}, a new review for a business item also in the field of \textit{\textbf{<label>}} with rating \textit{<attribute>} star(s) but diverse in the expression compared to the above given samples is: '' & \multirow{2}{*}{\shortstack{\textit{\textbf{Arts \& Entertainment /}}\\\textit{\textbf{Bars / Beauty \& Spas /}}\\\textit{\textbf{Event Planning \& Services /}}\\\textit{\textbf{Grocery / Health \& Medical /}}\\\textit{\textbf{Home \& Garden /}}\\\textit{\textbf{Hotels \& Travel /}}\\\textit{\textbf{Restaurants / Shopping}}}} & \multirow{2}{*}{\shortstack{1.0 / 2.0 / 3.0 /\\4.0 / 5.0}} \\
        \cline{2-3}
        \\[-1em]
        ~ & w/ Contrastive& A bad business review is: \textit{<sample\_1>}\emph{\textbackslash n}...\emph{\textbackslash n}A bad business review is: \textit{<sample\_$\lfloor$S/2$\rfloor$>}\emph{\textbackslash n}A good business review is: \textit{<sample\_$\lfloor$S/2$\rfloor$+1>}\emph{\textbackslash n}...\emph{\textbackslash n}A good business review is: \textit{<sample\_S}\emph{\textbackslash n}\emph{\textbackslash n}Based on the above examples of bad and good business reviews belonging to the category of \textit{\textbf{<label>}}, analyze the differences between the bad and good reviews. Generate a new review for a business item also in the field of \textit{\textbf{<label>}} with rating \textit{<attribute>} star(s) but diverse in the expression compared to the given good reviews. Ensure that the new review is further refined than the good reviews while maintaining clarity, making the good reviews appear to lie midway between the new review and the bad reviews. The new business review in the field of \textit{\textbf{<label>}} is: '' & ~ & ~ \\
        \hline
        \\[-1em]
        \multirow{2}{*}{\shortstack{Yelp-Rating\\(rating classification\\of business review)}} & w/o Contrastive & The business review is: \textit{<sample\_1>}\emph{\textbackslash n}The business review is: \textit{<sample\_2>}\emph{\textbackslash n}...\emph{\textbackslash n}The business review is: \textit{<sample\_S>}\emph{\textbackslash n}\emph{\textbackslash n}Based on the above business reviews with rating \textit{\textbf{<label>}} star(s), a new review for a business item in the field of \textit{<attribute>} also with rating \textit{\textbf{<label>}} star(s) but diverse in the expression compared to the above given samples is: '' & \multirow{2}{*}{\shortstack{\textit{\textbf{1.0 / 2.0 / 3.0 /}}\\\textit{\textbf{4.0 / 5.0}}}} & \multirow{2}{*}{\shortstack{Arts \&\\Entertainment /\\Bars /\\Beauty \& Spas /\\Event Planning \& Services /\\Grocery /\\Health \& Medical /\\Home \& Garden /\\Hotels \& Travel /\\Restaurants /\\Shopping}} \\
        \cline{2-3}
        \\[-1em]
        ~ & w/ Contrastive& A bad business review is: \textit{<sample\_1>}\emph{\textbackslash n}...\emph{\textbackslash n}A bad business review is: \textit{<sample\_$\lfloor$S/2$\rfloor$>}\emph{\textbackslash n}A good business review is: \textit{<sample\_$\lfloor$S/2$\rfloor$+1>}\emph{\textbackslash n}...\emph{\textbackslash n}A good business review is: \textit{<sample\_S}\emph{\textbackslash n}\emph{\textbackslash n}Based on the above examples of bad and good business reviews with rating \textit{\textbf{<label>}} star(s), analyze the differences between the bad and good reviews. Generate a new review for a business item in the field of \textit{<attribute>} also with rating \textit{\textbf{<label>}} star(s) but diverse in the expression compared to the above given good reviews. Ensure that the new review is further refined than the good reviews while maintaining clarity, making the good reviews appear to lie midway between the new review and the bad reviews. The new business review with rating \textit{\textbf{<label>}} star(s) is: '' & ~ & ~ \\
       \hline
        \\[-1em]
        \multirow{2}{*}{\shortstack{Openreview-Category\\(field classification\\of paper review)}} & w/o Contrastive & The paper review is: \textit{<sample\_1>}\emph{\textbackslash n}The paper review is: \textit{<sample\_2>}\emph{\textbackslash n}...\emph{\textbackslash n}The paper review is: \textit{<sample\_S>}\emph{\textbackslash n}\emph{\textbackslash n}Based on the above paper reviews of paper in the area \textit{\textbf{<label>}}, a new review for a paper also in the area of \textit{\textbf{<label>}} with final recommendation: '\textit{<attribute>}' but diverse in the expression compared to the above given samples is: '' & \multirow{2}{*}{\shortstack{\textit{\textbf{Applications / Deep Learning}}\\\textit{\textbf{and representational learning /}}\\\textit{\textbf{General Machine Learning /}}\\\textit{\textbf{Generative models /}}\\\textit{\textbf{Machine Learning for}}\\\textit{\textbf{Sciences / Neuroscience}}\\\textit{\textbf{and Cognitive Science /}}\\\textit{\textbf{Optimization / Probabilistic}}\\\textit{\textbf{Methods / Reinforcement}}\\\textit{\textbf{Learning / Social Aspects}}\\\textit{\textbf{of Machine Learning /}}\\\textit{\textbf{Theory / Unsupervised}}\\\textit{\textbf{and Self-supervised learning}}}} & \multirow{2}{*}{\shortstack{1: strong reject /\\ 3: reject, not good enough /\\ 5: marginally below the\\acceptance threshold /\\ 6: marginally above the\\acceptance threshold /\\ 8: accept, good paper}} \\
        \cline{2-3}
        \\[-1em]
        ~ & w/ Contrastive& A bad paper review is: \textit{<sample\_1>}\emph{\textbackslash n}...\emph{\textbackslash n}A bad paper review is: \textit{<sample\_$\lfloor$S/2$\rfloor$>}\emph{\textbackslash n}A good paper review is: \textit{<sample\_$\lfloor$S/2$\rfloor$+1>}\emph{\textbackslash n}...\emph{\textbackslash n}A good paper review is: \textit{<sample\_S}\emph{\textbackslash n}\emph{\textbackslash n}Based on the above examples of bad and good paper reviews of paper in the area \textit{\textbf{<label>}}, analyze the differences between the bad and good reviews. Generate a new review for a paper also in the area of \textit{\textbf{<label>}} with final recommendation: '\textit{<attribute>}' but diverse in the expression compared to the given good reviews. Ensure that the new review is further refined than the good reviews while maintaining clarity, making the good reviews appear to lie midway between the new review and the bad reviews. The new paper review in the area \textit{\textbf{<label>}} is: '' & ~ & ~ \\
        \hline
        \\[-1em]
        \multirow{2}{*}{\shortstack{Openreview-Rating\\(rating classification\\of paper review)}} & w/o Contrastive & The paper review is: \textit{<sample\_1>}\emph{\textbackslash n}The paper review is: \textit{<sample\_2>}\emph{\textbackslash n}...\emph{\textbackslash n}The paper review is: \textit{<sample\_S>}\emph{\textbackslash n}\emph{\textbackslash n}Based on the above paper reviews of final recommendation: \textit{\textbf{<label>}}, a new review for a paper in the field of '\textit{<attribute>}' also with final recommendation: \textit{\textbf{<label>}} but diverse in the expression compared to the above given samples is: '' & \multirow{2}{*}{\shortstack{\textit{\textbf{1: strong reject /}}\\\textit{\textbf{3: reject, not good enough /}}\\\textit{\textbf{5: marginally below the}}\\\textit{\textbf{acceptance threshold /}}\\\textit{\textbf{6: marginally above the}}\\\textit{\textbf{acceptance threshold /}}\\\textit{\textbf{8: accept, good paper}}}} & \multirow{2}{*}{\shortstack{Applications / Deep Learning\\and representational learning /\\General Machine Learning /\\Generative models /\\Machine Learning for\\Sciences / Neuroscience\\and Cognitive Science /\\Optimization / Probabilistic Methods /\\Reinforcement Learning /\\Social Aspects of\\Machine Learning /\\Theory / Unsupervised\\and Self-supervised learning}} \\
        \cline{2-3}
        \\[-1em]
        ~ & w/ Contrastive& A bad paper review is: \textit{<sample\_1>}\emph{\textbackslash n}...\emph{\textbackslash n}A bad paper review is: \textit{<sample\_$\lfloor$S/2$\rfloor$>}\emph{\textbackslash n}A good paper review is: \textit{<sample\_$\lfloor$S/2$\rfloor$+1>}\emph{\textbackslash n}...\emph{\textbackslash n}A good paper review is: \textit{<sample\_S}\emph{\textbackslash n}\emph{\textbackslash n}Based on the above examples of bad and good paper reviews of final recommendation: \textit{\textbf{<label>}}, analyze the differences between the bad and good reviews. Generate a new review for a paper in the field of '\textit{<attribute>}' also with final recommendation: \textit{\textbf{<label>}} but diverse in the expression compared to the above given good reviews. Ensure that the new review is further refined than the good reviews while maintaining clarity, making the good reviews appear to lie midway between the new review and the bad reviews. The new paper review of final recommendation: \textit{\textbf{<label>}} is: '' & ~ & ~ \\
        \hline
        \\[-1em]
        \multirow{2}{*}{\shortstack{Banking\\(field classification\\of online banking\\queries)}} & w/o Contrastive & The online banking query is: \textit{<sample\_1>}\emph{\textbackslash n}The online banking query is: \textit{<sample\_2>}\emph{\textbackslash n}...\emph{\textbackslash n}The online banking query is: \textit{<sample\_S>}\emph{\textbackslash n}\emph{\textbackslash n}Based on the above online banking queries in the category of ``\textit{\textbf{<label>}}'', a new online banking query also in the category of ``\textit{\textbf{<label>}}'' but diverse in the expression compared to the above given samples is: '' & \multirow{2}{*}{\shortstack{\textit{\textbf{activate\_my\_card }}/\\\textit{\textbf{age\_limit /}}\\\textit{\textbf{apple\_pay\_or\_google\_}}\\\textit{\textbf{pay / atm\_support /}}\\\textit{\textbf{automatic\_top\_up /}}\\\textit{\textbf{balance\_not\_updated\_}}\\\textit{\textbf{after\_bank\_transfer /}}\\\textit{\textbf{ balance\_not\_updated\_}}\\\textit{\textbf{after\_cheque\_or\_cash\_}}\\\textit{\textbf{deposit / beneficiary\_}}\\\textit{\textbf{not\_allowed / cancel\_}}\\\textit{\textbf{transfer / card\_}}\\\textit{\textbf{about\_to\_expire}}}} & \multirow{2}{*}{None} \\
        \cline{2-3}
        \\[-1em]
        ~ & w/ Contrastive& A bad online banking query is: \textit{<sample\_1>}\emph{\textbackslash n}...\emph{\textbackslash n}A bad online banking query is: \textit{<sample\_$\lfloor$S/2$\rfloor$>}\emph{\textbackslash n}A good online banking query is: \textit{<sample\_$\lfloor$S/2$\rfloor$+1>}\emph{\textbackslash n}...\emph{\textbackslash n}A good online banking query is: \textit{<sample\_S}\emph{\textbackslash n}\emph{\textbackslash n}Based on the above examples of bad and good online banking queries in the category of ``\textit{\textbf{<label>}}'', analyze the differences between the bad and good reviews. Generate a new online banking query also in the category of ``\textit{\textbf{<label>}}'' but diverse in the expression compared to the above given good queries. Ensure that the new query is further refined than the good queries while maintaining clarity, making the good queries appear to lie midway between the new query and the bad queries. The new online banking query also in the category of ``\textit{\textbf{<label>}}'' is: '' & ~ & ~ \\
    \bottomrule
    \end{tabular}
    }
\end{table*}
In \cref{tab:appendix_prompt}, we listed the prompts used in our experiments, including contrastive (``w Contrastive'') and non-contrastive (``w/o Contrastive'') in-context learning prompts. We need to clarify that, for PE series baselines, we use their original prompt for \verb|VARIATIONAL_API|, which is different from the listed ``w/o Contrastive in-context learning'' prompt in \cref{tab:appendix_prompt}. Please refer to \citet{xie2024differentially} (the original work) for detailed prompts.

\section{Algorithm for Distributed Private Data and Detailed Functions} \label{sec:appendix_algorithms}

Due to space limitation, we include the full algorithm for $L>1$ setting here in \cref{alg:algorithm_full_functions} in the Appendix. The difference between \cref{alg:algorithm_full_functions} and \cref{alg:algorithm_full_functions_singlePDP} mainly falls in line 2 and lines 8 to 11 in \cref{alg:algorithm_full_functions}.

\begin{algorithm}[tb]
\caption{WASP for Distributed Federated Data ($L>1$)} 
\label{alg:algorithm_full_functions}
\begin{flushleft}
\textbf{Input:}\\ \quad
$K$ PLMs $\{\mathcal{P}_{k}\}_{k=1}^K$ with empty synthetic dataset $\{\mathcal{D}_k\leftarrow\emptyset\}_{k=1}^K$;\\ \quad
$L$ private data parties controlling distributed private dataset $\{\mathcal{B}_l\}_{l=1}^L$ of $M$ samples in total that belongs to $C$ categories;\\ \quad
number of in-context samples $S$;\\ \quad
number of iterations $T$ taken to obtain in total $N$ synthetic samples;\\ \quad
initialized PLM weights ${\{w_{k}=1/K\}}_{k=1}^K$;\\ \quad
learning rate $\eta$;\\ \quad
DP privacy parameters $\epsilon,\delta$;\\ \quad
test dataset of downstream task $\mathcal{A}$;\\ \quad
random initialized STM $m_{(0)}$; 
\\
\textbf{Output:} STM $m$. 
\end{flushleft}

\begin{algorithmic}[1]
    \STATE Initialize in-context feedback samples $\hat{\mathcal{D}}^n \leftarrow \emptyset, \hat{\mathcal{D}}^f \leftarrow \emptyset$.
    \STATE Calculate Gaussian noise $\sigma=4\frac{\sqrt{2\log{\left(1.25/\delta\right)}}\sqrt{T-1}}{\epsilon\sqrt{L}}$.
    \FOR{$t=0$ {\bfseries to} $T-1$}
        \FOR{$k=1$ {\bfseries to} $K$ {\bfseries in parallel}}
            \STATE $\mathcal{D}_k \leftarrow$ \verb|WeightedSynDataGeneration(|$\mathcal{D}_k$, $\hat{\mathcal{D}}^n$, $\hat{\mathcal{D}}^f$, $\left[ (N/T)\times w_{k} \right]$, $C$\verb|)|.
        \ENDFOR
        \STATE $\mathcal{D} \leftarrow \cup_{k=1}^{K}\mathcal{D}_k$.
        \FOR{$l=1$ {\bfseries to} $L$ {\bfseries in parallel}}
            \STATE $H^{n}_l, H^{f}_l \leftarrow$ \verb|DP_PrivateVoting(|$\mathcal{D}$, $\mathcal{B}_l$, $Q$, $\sigma$\verb|)|.
        \ENDFOR
        \STATE $H^{n}\leftarrow\sum_{l=1}^{L} H^{n}_l$; $H^{f}\leftarrow\sum_{l=1}^{L} H^{f}_l$.
        \STATE $\hat{\mathcal{D}}^n, \hat{\mathcal{D}}^f \leftarrow$ \verb|SampleSelection(|$\mathcal{D}$, $H^{n}$, $H^{f}$, $S$, $C$\verb|)|.
        \STATE $\{w_k\}_{k=1}^K \leftarrow$ \verb|PLMScoring(|$H^{n}$, $\{\mathcal{D}_k\}_{k=1}^K$\verb|)|.
    \ENDFOR
    \STATE $m \leftarrow$ \verb|STMTraining(|$\mathcal{D}$, $m_{(0)}$, $\eta$\verb|)|.
\end{algorithmic}
\end{algorithm}

We also include pseudo-code for the functions used in \cref{alg:algorithm_full_functions_singlePDP,alg:algorithm_full_functions} here in \cref{alg:algorithm_functions} due to space limitation.

\begin{algorithm*}[!htb]
\small
\caption{Functions used in \cref{alg:algorithm_full_functions_singlePDP,alg:algorithm_full_functions} for WASP}
\label{alg:algorithm_functions}
    \begin{flushleft}\textbf{function} \verb|WeightedSynDataGeneration(|$\mathcal{D}_k$, $\hat{\mathcal{D}}^n$, $\hat{\mathcal{D}}^f$, $\hat{N}$, $C$\verb|)|:\end{flushleft}
    
\begin{algorithmic}[]
    \FOR{$c=1$ {\bfseries to} $C$}
        \IF{$t=0$}
            \STATE Use zero-shot prompt as working prompt $\mathcal{T}(c)$.
        \ELSE
            \STATE Randomly sample $S-\lfloor S/2 \rfloor$ samples from $\hat{\mathcal{D}}^{n,[c]}$ and $\lfloor S/2 \rfloor$ samples from $\hat{\mathcal{D}}^{f,[c]}$ to create few-shot prompt as working prompt $\mathcal{T}(c)$.
        \ENDIF
        \STATE Generate $\lceil\hat{N}/C\rceil$ samples using $\mathcal{T}$ and add them to $\mathcal{D}_k$.
    \ENDFOR
    \IF{$|\mathcal{D}_k|>\hat{N}$}
        \STATE Random sample $|\mathcal{D}_k|-\hat{N}$ different samples from $\mathcal{D}_k$ and remove them from $\mathcal{D}_k$.
    \ENDIF
    \STATE \textbf{return} $\mathcal{D}_k$.

    \STATE
\end{algorithmic}

    \begin{flushleft}\textbf{function} \verb|STMTraining(|$\mathcal{D}$, $m_{(0)}$, $\eta$\verb|)|:\end{flushleft}
    
\begin{algorithmic}[]
    \STATE Initialize a trainable STM ${m} \leftarrow m_{(0)}$.
    \STATE Train ${m}$ using $\mathcal{D}$ with learning rate $\eta$ till convergence by using objective function $\mathcal{L} = \sum_{i=1}^{|\mathcal{D}|} \ell(m(\mathbf{x}_{i}),y_{i})$.
    \STATE \textbf{return} $m$.

    \STATE
\end{algorithmic}

    \begin{flushleft}\textbf{function} \verb|DP_PrivateVoting(|$\mathcal{D}$, $\mathcal{B}$, $Q$, $\sigma$\verb|)|:\end{flushleft}

\begin{algorithmic}[]
    \STATE Initialized $H^{n}\leftarrow[0,\dots,0]$; $H^{f}\leftarrow[0,\dots,0]$ of length $\mathbb{R}^{|\mathcal{D}|}$ and note the total DP synthetic dataset as $\mathcal{D}=\{(\textbf{x}_i, y_i)\}_{i=1}^{|\mathcal{D}|}$.
    \FOR{($\textbf{z}_j, u_j$) {\bfseries in} $\mathcal{B}$}
        \STATE $\mathcal{D}^{[u_j]} = \left\{ (\mathbf{x}_i,y_i) \in \mathcal{D} \,|\, y_i=u_j \right\}$.
        \STATE $[n_{j,1},\dots,n_{j,Q}] \leftarrow \arg\mathrm{top}Q\mathrm{Smallest} \left( \, d(\textbf{z}_j, \textbf{x}_{i})_{(\mathbf{x}_{i},y_i)\in \mathcal{D}^{[u_j]}} \right)$.
        \STATE $[f_{j,1},\dots,f_{j,Q}] \leftarrow \arg\mathrm{top}Q\mathrm{Largest} \left( \, d(\textbf{z}_j, \textbf{x}_{i})_{(\mathbf{x}_{i},y_i)\in \mathcal{D}^{[u_j]}} \right)$.
        \FOR{$q=1$ {\bfseries to} $Q$}
            \STATE $H^{n}[n_{j,q}] \leftarrow H^{n}[n_{j,q}]+\frac{1}{2^{q-1}}$.
            \STATE $H^{f}[f_{j,q}] \leftarrow H^{f}[f_{j,q}]+\frac{1}{2^{q-1}}$.
        \ENDFOR
    \ENDFOR
    \STATE $H^{n} \leftarrow H^{n} + \mathcal{N}(0,\sigma^2I_{|\mathcal{D}|})$.
    \STATE $H^{f} \leftarrow H^{f} + \mathcal{N}(0,\sigma^2I_{|\mathcal{D}|})$.
    \STATE \textbf{return} $H^{n}$, $H^{f}$.

    \STATE
\end{algorithmic}




    \begin{flushleft}\textbf{function} \verb|PLMScoring(|$H$, $\{\mathcal{D}_k\}_{k=1}^K$\verb|)|:\end{flushleft}

\begin{algorithmic}[]
    \FOR{$k=1$ {\bfseries to} $K$}
        \STATE Calculate $s_{i} = H^n[i] \,\big/ \sum_{i'=1}^{|\mathcal{D}|} H^n[i']$.
        \STATE Calculate model score $w_k = \frac{\sum_{(\textbf{x}_{i},y_{i})\in\mathcal{D}_k} s_{i}}{|\mathcal{D}_{k}| / \sum_{k'=1}^K |\mathcal{D}_{k'}|}= \frac{\sum_{(\textbf{x}_{i},y_{i})\in\mathcal{D}_k} s_{i}}{|\mathcal{D}_{k}| / |\mathcal{D}|}$
    \ENDFOR
    \STATE \textbf{return} $\{w_k\}_{k=1}^K$.

    \STATE
\end{algorithmic}

    \begin{flushleft}\textbf{function} \verb|SampleSelection(|$\mathcal{D}$, $H^{n}$, $H^{f}$, $S$, $C$\verb|)|:\end{flushleft}
    
\begin{algorithmic}[]
    \STATE Reset $\hat{\mathcal{D}^n} \leftarrow \emptyset$, $\hat{\mathcal{D}^f} \leftarrow \emptyset$.
    \STATE $\mathcal{H}^n \leftarrow \frac{H^n}{\sum_{i=1}^{|\mathcal{D}|} H^n[i]}$.
    \STATE $\mathcal{H}^f \leftarrow \frac{H^f}{\sum_{i=1}^{|\mathcal{D}|} H^f[i]}$.
    \FOR{$c=1$ {\bfseries to} $C$}
        \STATE $\mathcal{D}^{[c]} = \left\{ (\mathbf{x}_i,y_i) \in \mathcal{D} \,|\, y_i=c \right\}$.
        \STATE $H^{n,[c]} = \left\{ H^n[i] \,\big|\, (\mathbf{x}_{i},y_i) \in \mathcal{D}^{[c]} \right\}, H^{f,[c]} = \left\{ H^f[i] \,\big|\, (\mathbf{x}_{i},y_i) \in \mathcal{D}^{[c]} \right\}$.
        \STATE $\hat{\mathcal{D}}^{n,[c]} = \left\{ (\mathbf{x}_i,y_i) \in \mathcal{D}^{[c]} \,\big|\, H^n[i] \text{ is among the top-}S \text{ values of } H^{n,[c]} \right\}$, contains the top-$S$ samples from $\mathcal{D}^{[c]}$ ranked by $\mathcal{H}^{n,[c]}$.
        \STATE $\hat{\mathcal{D}}^{f,[c]} = \left\{ (\mathbf{x}_i,y_i) \in \mathcal{D}^{[c]} \,\big|\, H^f[i] \text{ is among the top-}S \text{ values of } H^{f,[c]} \right\}$, contains the top-$S$ samples from $\mathcal{D}^{[c]}$ ranked by $\mathcal{H}^{f,[c]}$.
    \ENDFOR
    \STATE $\hat{\mathcal{D}}^n=\left\{\hat{\mathcal{D}}^{n,[1]},\dots,\hat{\mathcal{D}}^{n,[C]}\right\}$, $\hat{\mathcal{D}}^f=\left\{\hat{\mathcal{D}}^{f,[1]},\dots,\hat{\mathcal{D}}^{f,[C]}\right\}$.
    \STATE \textbf{return} $\hat{\mathcal{D}}^n, \hat{\mathcal{D}}^f$.

\end{algorithmic}
\end{algorithm*}

\section{Supporting Results for Introduction} \label{sec:appendix_intro_support}

\subsection{Examples of High-quality and Low-quality Samples} \label{subsec:appendix_good_bad_samples}
We show examples of high-quality and low-quality synthetic samples generated using Aug-PE in \cref{tab:examples_good_samples} and \cref{tab:examples_bad_samples} respectively. We also include the appearance frequency of some types of low-quality samples within the generated DP synthetic dataset in \cref{tab:examples_bad_samples}.

\cref{tab:examples_bad_samples} shows that, low-quality noisy samples often diverge from the specified task (generating movie reviews in positive/negative sentiment for this table). Differently, likes shown in \cref{tab:examples_good_samples}, high-quality samples often possess a clear sentiment tendency that well accomplished the task, with some offering detailed judgments or even containing concession details.

\begin{table*}[]
\caption{Low-quality DP synthetic samples for movie review semantic analysis with IMDb as real dataset.}
\label{tab:examples_bad_samples}
\begin{small}
    \centering
    \resizebox{1\linewidth}{!}{
    \begin{tabular}{l|p{8.5cm}|c|p{4.5cm}}
    \toprule
        \textbf{Model} & \textbf{Low-quality Noisy Sample Text (Examples)} & \textbf{Label} & \textbf{Explain} \\
    \midrule
        GPT-2 & ``\textasciitilde \ If you are missing No. 17 (see below) \textasciitilde '' & negative & Meaning less sentence.\\
        \hline
        \\[-1em]
        Llama-2 & ``In an informal way, please rephrase the sentences as follows:'' & positive & Repeating the prompt. Appears around $30$ times in $\mathcal{D}$ of size $6,000$.\\
        \hline
        \\[-1em]
        Vicuna & {``\begin{CJK}{UTF8}{} \CJKfamily{mj} \textbackslash n번역결과  \textbackslash n좋은 감정을 기반으로 영화 관람 후 즐거움과 엔터테인먼트 가치를 즐기셨기에 좋습니다. \end{CJK}''} & positive & Unmatched language and translating the prompt. Appears within $9$ samples in $\mathcal{D}$ of size $6,000$.\\
        \hline
        \\[-1em]
        OPT & ``The sentence is positive, but is ambiguous. Would it make sense to remove '' & negative & Task unrelated and mislabeled.\\
        \hline
        \\[-1em]
        ChatGLM3 & ``... {\begin{CJK}{UTF8}{gkai}指导意见：句子间不顺畅，需要调整一下顺序，去掉一些冗余的词汇。\end{CJK}} Since is based on positive sentiment, then how does one use the 'pronoun?'' & positive & Unmatched language and repeating the prompt.\\
        \hline
        \\[-1em]
        Flan-T5 & ``what a fantastic movie!! :O great acting and spectacular scenery. great work. i look forward to another good one next year :D come on laura delana!!!!!!!!!!!!!!!!!!!!!!!!!!!!!!!!!!!!!!!!!!!!!!!! woohoooo!!!....can't wait!!!!!!!!!!!!!!!!!!!!!!!!!!!!!!!!!!!!!!!!!!!!!!!!!!!!!!!!!!!!!!!!!!!!!!!!!!....'' & positive & Excessively long consecutive punctuation marks. Appears within $4$ samples in $\mathcal{D}$ of size $6,000$.\\
    \bottomrule
    \end{tabular}
    }
\end{small}
\end{table*}

\begin{table*}[]
\caption{High-quality DP synthetic samples for movie review semantic analysis with IMDb as real dataset.}
\label{tab:examples_good_samples}
\begin{small}
    \centering
    \resizebox{1\linewidth}{!}{
    \begin{tabular}{l|p{13.5cm}|c}
    \toprule
        \textbf{Model} & \textbf{High-quality Sample Text (Examples)} & \textbf{Label}\\
    \midrule
        GPT-2 & ``This movie isn't for everyone, but it's worth a look. But if you have to be in the dark to enjoy it, just buy the soundtrack.'' & positive \\
        \hline
        \\[-1em]
        GPT-2 & ``is that it has some semi-interesting world-building, but the information it offers is grossly insufficient to allow one to progress in the story. The central characters have little to do with the world beyond playing around with their toys and talking to themselves. '' & negative \\
        \hline
        \\[-1em]
        Llama-2 & ``Incredible film! I laughed, I cried, and I left the theater feeling inspired. Every performance was top-notch and the direction was flawless. A true masterpiece that will be remembered for years to come. 10/10.'' & positive \\
        \hline
        \\[-1em]
        Llama-2 & ``This film was a colossal disappointment. I found myself glancing at my wrist every 10 minutes, eagerly anticipating its conclusion. The narrative was intricate and challenging to comprehend, and the personas were underdeveloped. I had no emotional investment in their plight. The performances were lacking, and the dialogue was trite and unconvincing. In summary, I would not advise viewing this motion picture to any individual. '' & negative \\
        \hline
        \\[-1em]
        Vicuna & ``A movie for people who want to see an action-packed adventure that will keep them on the edge of their seats.'' & positive \\
        \hline
        \\[-1em]
        Vicuna & ``Felt like a missed opportunity. Glacial pacing and shallow characters. The premise was intriguing but the execution fell flat. Avoid this one.'' & negative \\
        \hline
        \\[-1em]
        OPT & ``-This film is enjoyable to watch because of it's well-developed storyline and most of all its storyline that i can relate to and would like to have. This film also has good acting performances and the songs help the film a lot. It's not a typical love story kind of a film, but it does have lots of interesting messages to give. Very well-done! Definitely recommend this film!'' & positive \\
        \hline
        \\[-1em]
        OPT & ``Quote: The first sentence means nothing to me. Without context the first two sentences also mean nothing.'' & negative \\
        \hline
        \\[-1em]
        ChatGLM3 & ``Attention getter, visually interesting and outstanding acting, the story of an American citizen that is in Mexico and gets involved in a murder is a good movie. '' & positive \\
        \hline
        \\[-1em]
        ChatGLM3 & ``This movie is an unwatchable mess, basically a rip off of that scene in Dr. Strangelove where the U.S. Air Force gets into a scramble to deliver nukes after a false alert... I like pretty much every aspect of anime, but this stuff is bad. Animation is choppy, sound effects and voice acting are atrocious, and the plot makes little sense.'' & negative \\
        \hline
        \\[-1em]
        Flan-T5 & ``<pad> This is a fine historical drama about America's Civil Rights movement, that has William Friedkin as director and Gary Cooper as an actor. Although there are other excellent directors that we haven't heard of, William Friedkin has become the standard in many ways. Filming in New Orleans, Mr. Friedkin gives a glimpse into the African-American experience during the late 1920s. I saw it at the All-African Cinema festival at Wharton Business School. I was awestruck by the power and impact of the film. {\color{gray}{\textit{...[truncated because of excessive length]...}}}'' & positive \\
        \hline
        \\[-1em]
        Flan-T5 & ``<pad> beware of this film! i can't recall a single horror movie that was scarier than this one. it's just plain cheesy and stupid. there's no plot and the characters are entirely cheesy. so don't even bother. you'll never forget it.'' & negative \\
    \bottomrule
    \end{tabular}
    }
\end{small}
\end{table*}

\section{Theoretical Privacy Analysis for WASP} \label{sec:appendix_privacy_analysis}

To prove \cref{theorem:dp_L_equals_1}, in this part, we prove that the WASP framework described in \cref{alg:algorithm_full_functions} with distributed federated data ($L>1$) satisfies $(\epsilon,\delta)$-DP, which is the general case for $L=1$ setting described in \cref{alg:algorithm_full_functions_singlePDP} and \cref{theorem:dp_L_equals_1}. 

\begin{theorem} \label{theorem:gaussian_dp}
    Let $f$ be a function with global $L_2$ sensitivity $\Delta$. For any $\epsilon>0,\delta \in (0,1)$, the Gaussian output perturbation mechanism with $\sigma=\Delta\frac{\sqrt{2\log\left({1.25/\delta}\right)}}{\epsilon}$ ensures that $f$ satisfies $(\epsilon,\delta)$-DP.
\end{theorem}
Proof of \cref{theorem:gaussian_dp} can be found in \citet{balle2018improving}.

\begin{theorem} \label{theorem:function_sensitivity}
    The global $L_2$ sensitivity $\Delta$ of WASP described in \cref{alg:algorithm_full_functions} is $4$.
\end{theorem}
\begin{proof}[Proof]
    In WASP (\cref{alg:algorithm_full_functions}), function \verb|DP_PrivateVoting| is the only function that accesses the private dataset $\mathcal{B}$. Thus, $\Delta$ of WASP equals to that of function \verb|DP_PrivateVoting|. Within function \verb|DP_PrivateVoting|, for nearest histogram and furthest histogram respectively, each private sample contributes $Q$ votes with decaying voting weights $\{1,\frac{1}{2},\dots,\frac{1}{2^{Q-1}}\}$. Therefore, the total votes contributed by one private sample is $\sum_{q=1}^{Q}\frac{1}{2^{q-1}}=2-\frac{1}{2^{Q-1}}<2$ for each histogram. Adding or removing one private sample in $\mathcal{B}$ will result in a change no more than $2$ in the $\ell_2$ norm of each histogram. Therefore, the upper bound of the sensitivity for each histogram is $2$ and the upper bound of the sensitivity of WASP framework is $4$ considering both histograms, i.e. $\Delta=4$. 
\end{proof}

\begin{lemma} \label{lemma:T_iterations}
    If a Gaussian mechanism satisfies $(\epsilon,\delta)$-DP, then independently repeating this mechanism for $T$ times results in the DP budget to increase to $\sqrt{T}\cdot\epsilon$.
\end{lemma}
The proof of \cref{lemma:T_iterations} can be found in \citet{steinke2022composition}.

With the above lemma and theorems, we present and prove our main theorem as follows.
\begin{theorem} 
\label{theorem:dp}
    If each private data party performs standard Gaussian mechanism with addition noise following $\mathcal{N}(0,\sigma^2)$ and $\sigma=4\frac{\sqrt{2\log{\left(1.25/\delta\right)}}\sqrt{T-1}}{\epsilon\sqrt{L}}$, WASP framework described in \cref{alg:algorithm_full_functions} satisfies $(\epsilon,\delta)$-DP for private samples in $\mathcal{B}$.
\end{theorem}
\begin{proof}[Proof]
    For guaranteeing $(\epsilon,\delta)$-DP throughout the $T-1$ iterations with feedback (the first generation iteration does not use feedback), each iteration should satisfy a differential privacy budget of $\frac{\epsilon}{\sqrt{T-1}}$. Given $\Delta=4$ for WASP, $\sigma_{total}=4\frac{\sqrt{2\log{\left(1.25/\delta\right)}}\sqrt{T-1}}{\epsilon}$ for each single generation iteration will guarantee $(\epsilon,\delta)$-DP for the whole process. Further, Gaussian random variables satisfy that $X+Y\sim\mathcal{N}(0,\sigma_1^2+\sigma_2^2)$ if $X\sim\mathcal{N}(0,\sigma_1^2), Y\sim\mathcal{N}(0,\sigma_2^2)$ are independent. Therefore, if each private data party adds \textit{i.i.d.} Gaussian noise following $\mathcal{N}(0,\sigma^2)$ with $\sigma=4\frac{\sqrt{2\log{\left(1.25/\delta\right)}}\sqrt{T-1}}{\epsilon\sqrt{L}}$, the total noise follows $\mathcal{N}(0,\sigma_{total}^2)$ which guaranties $(\epsilon,\delta)$-DP for the whole WASP process.
\end{proof}

\section{Additional Results} \label{sec:appendix_results}

\subsection{Comparison of Synthetic Sample Resemblance for WASP} \label{subsec:appendix_fid_comparison}

\begin{figure}[tb]
    \centering
    \includegraphics[width=0.99\linewidth]{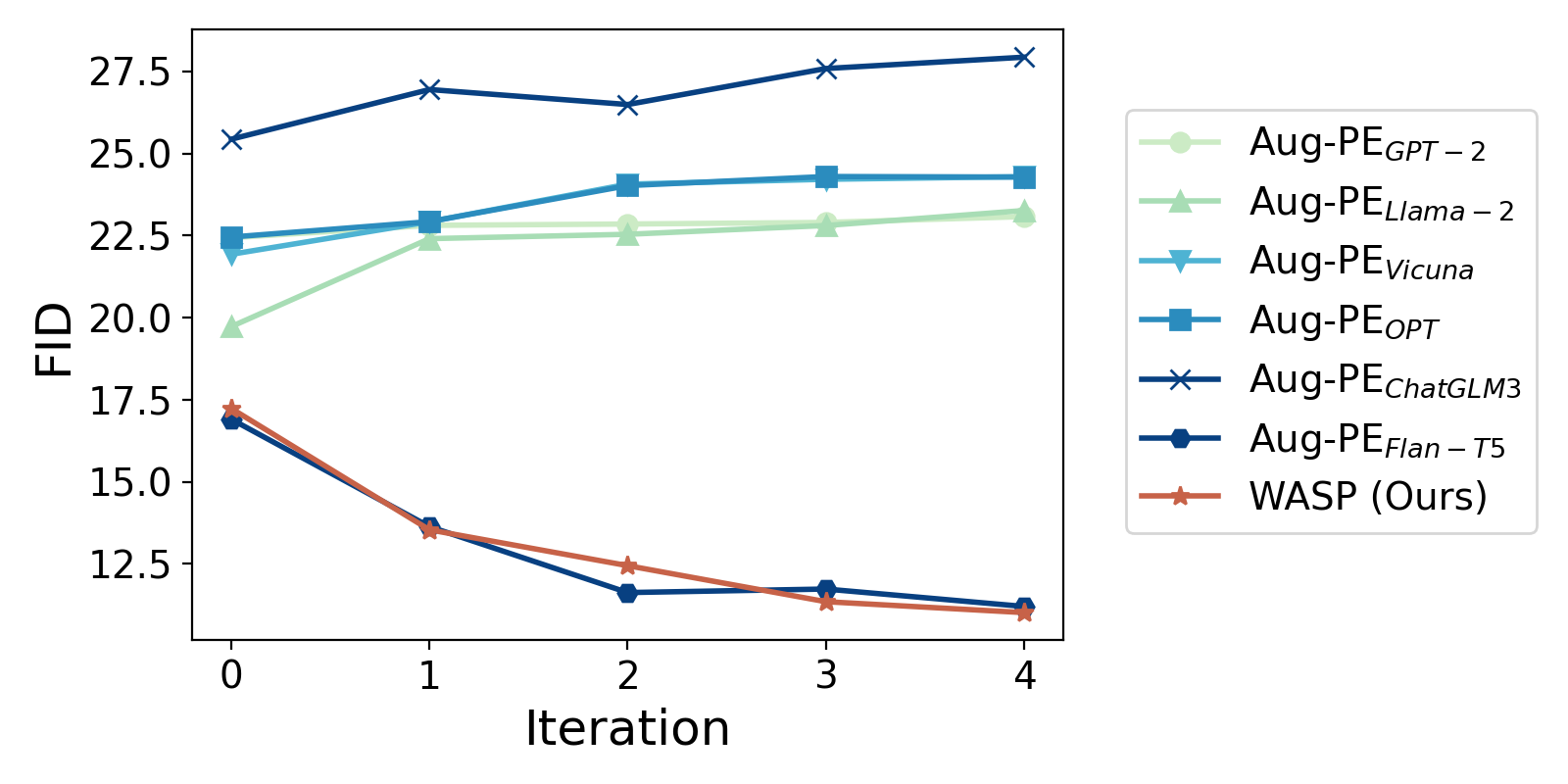}
    \caption{Comparison of the resemblance of synthetic dataset to real private dataset (FID) using Aug-PE and our proposed WASP using movie review semantic analysis task and IMDb dataset.}
    \label{fig:appendix_fid_comparison}
\end{figure}

To further demonstrate the effectiveness of WASP under limited private sample setting, we additionally use FID between the generated DP synthetic dataset $\mathcal{D}$ and the real private dataset $\mathcal{B}$ to evaluate the resemblance of the former ($\mathcal{D}$) to the later ($\mathcal{B}$) with $M=100$ in \cref{fig:appendix_fid_comparison}. Lower FID indicates higher distribution similarity therefore indicating better resemblance. 

As shown in \cref{fig:appendix_fid_comparison}, the baseline method Aug-PE often fails to generate a DP synthetic dataset that closely resembles $\mathcal{B}$ when using an improper PLM, leading to an increased FID value over iterations. On the contrary, WASP results in a consistently decreasing FID value over iteration, demonstrating it effectiveness in improving the resemblance of $\mathcal{D}$ to $\mathcal{B}$. Moreover, although WASP initially has a higher FID than using the best single PLM (Flan-T5 in \cref{fig:appendix_fid_comparison}) for Aug-PE (which is reasonable due to the initialization of $\mathcal{D}$ being a mixture of synthetic samples from different PLMs, making it better than the one generated solely by worst PLM but worse than the one given solely by the best PLM), it ultimately achieves a lower FID than all baseline counterparts. This indicates that our proposed WASP method better handles the limited private sample setting.

\subsection{Effectiveness of Differentially Private Top-$Q$ Voting and Contrast In-context Learning with Single PLM} \label{subsec:appendix_single_plm_topQ_contrast}

\begin{figure*}
    \centering
    \includegraphics[width=0.89\linewidth]{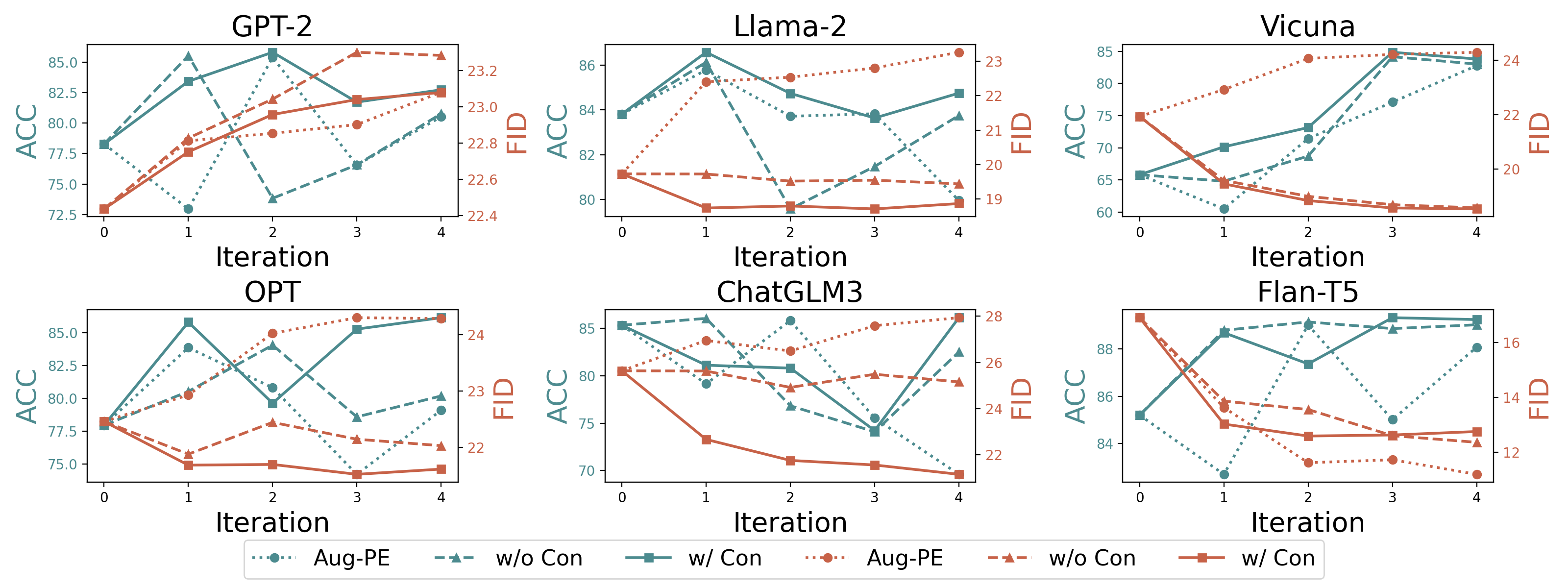}
    \caption{Comparison of the resemblance of synthetic dataset to real private dataset (Fréchet Inception Distance, FID) and trained downstream model performance (ACC) using Aug-PE (``Aug-PE'', doted lines), refinement on $Q=8$ without contrastive in-context learning (``w/o Con'', dashed lines) and refinement on $Q=8$ with contrastive in-context learning (``w/ Con'', solid lines) with single-PLM setting and $L=1$ under $(4.0, 1\times10^{-5})$-DP with IMDb dataset. 
    }
    \label{fig:pe_q8_contrast_comparison}
\end{figure*}

We present additional results to validate the effectiveness of \textit{Differentially Private Top-$Q$ voting} and \textit{contrastive in-context learning} with single PLM in \cref{fig:pe_q8_contrast_comparison}. Starting with Aug-PE, we increase $Q$ from $1$ to $8$ to obtain the ``w/o Con'' results, and then incorporate contrastive in-context learning samples into the prompt to achieve the ``w/ Con’’ results (also the $K=1$ setting for WASP). This refinement process shows a steady decrease in FID for 
most PLMs.
Nonetheless, an overall performance improvement is observed for all tested PLMs, both in terms of highest performance across iterations and final performance.

\subsection{Sensitive Analysis of $M$ for PE} \label{subsec:appendix_pe_gold_num_change}
\begin{figure*}[tb]
    \centering
    \includegraphics[width=0.89\linewidth]{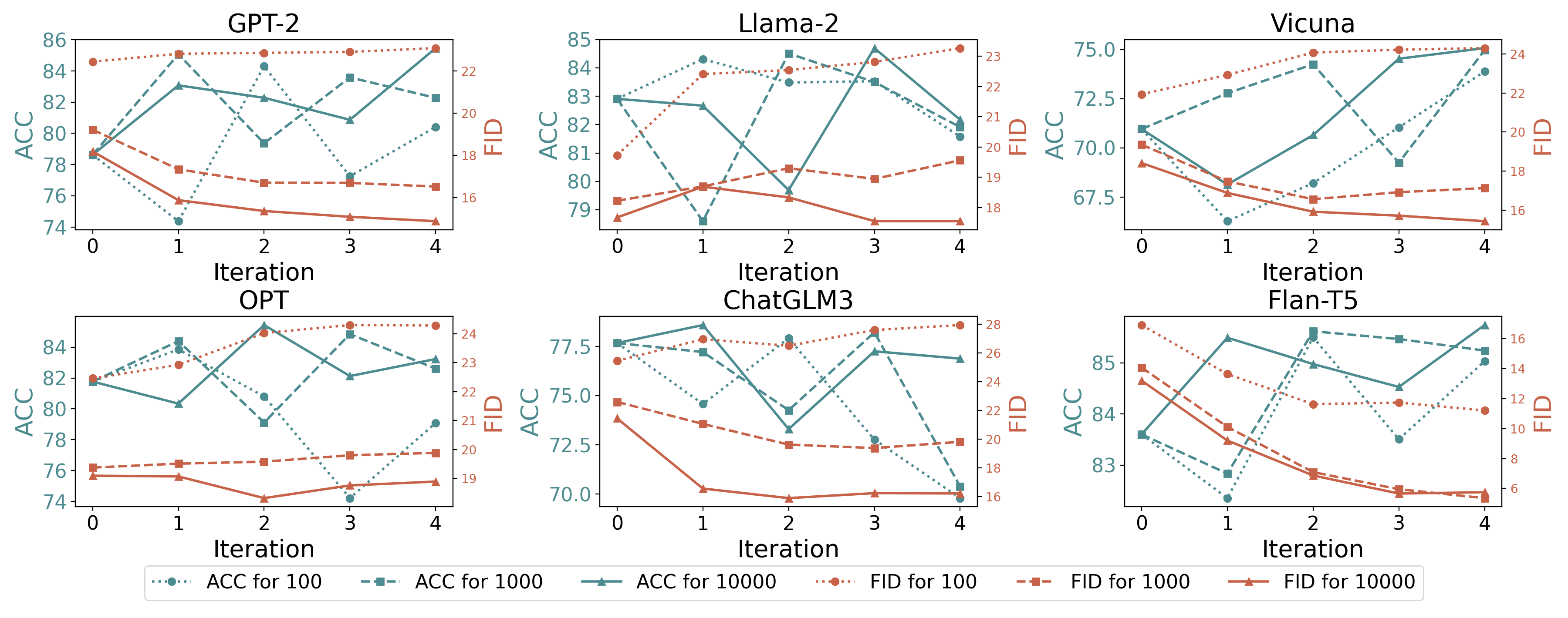}
    \caption{Comparison of the similarity of synthetic dataset to real private dataset (FID) and trained downstream model performance (ACC) with different amount of available private samples ($M$) using Aug-PE with $L=1$ under $(4.0, 1\times10^{-5})$-DP with IMDb dataset.}
    \label{fig:gold_change_fid_and_acc}
\end{figure*}

We performed experiments to analyze the sensitivity of Aug-PE~\cite{xie2024differentially} on various $M$ values. Results are included in \cref{fig:gold_change_fid_and_acc} which shows that most PLMs fail when only a limited amount of private samples ($M=100$) is available, with an increasing FID through iterations. Conversely, with sufficient amount of private samples ($M=10,000$), a continuous decrease in FID as well as less performance fluctuation can be observed throughout the iterations.

\subsection{Comparison of WASP and Pre-Text Under User-level DP} \label{subsec:appendix_user_level_dp_results}
In our work, we assume a full participation setting where all $L$ parties participate in each iteration. Based on this, we primarily focus on ensuring sample-level DP to protect each private sample $(\mathbf{z}_j,u_j) \in \mathcal{B}$ in this work.
However, our proposed WASP method can be easily extended to user-level DP protection and is also effective in protecting user-level DP compared to baselines (see \cref{tab:main_results_150}).

In \citet{hou2024pretext}, although they also study a full participation setting with $L>1$, they focus on user-level DP with the assumption that each participating private data party in the collaboration controls only a tiny amount of private samples ($8$ in their work). Therefore, following \citet{hou2024pretext}, to testify the effectiveness of WASP when extended to user-level DP, we assume that each participating data party controls no more than $8$ real private samples, i.e. $M_l \leq 8, l=1,\dots,L$. These distributed private datasets still aggregate to an unbalanced dataset like in \cref{subsec:experimental_settings}.

Under this setting, to protect user-level DP (where adding or removing one private data party should not significantly affect the function output), the function sensitively $\Delta_{user}$ should be $\max(M_1,$ $\dots,M_L)$ times as large as that for protecting sample-level DP ( $\Delta_{sample}$). The rational is that, the addition or removal of a private data party can result in the addition or removal of up to $\max(M_1,\dots,M_L)$ samples, leading to a change of no more than $\max(M_1,\dots,M_L) \times \Delta_{sample}$ in the $\ell_2$ distance of the produced histograms. Given that $\Delta_{sample}=4$ for WASP (see \cref{theorem:function_sensitivity} in \cref{sec:appendix_privacy_analysis} for details), we have $\Delta_{user}=\max(M_1,\dots,M_L) \times\Delta_{sample} \leq8\times\Delta_{sample}=8\times4=32$. In our experiments, we use $32$, the upper bound of $\Delta_{user}$, as the function sensitivity to calculate $\sigma=32\frac{\sqrt{2\log{\left(1.25/\delta\right)}}\sqrt{T-1}}{\epsilon\sqrt{L}}$ for $(\epsilon,\delta)$-DP protection.

Results are shown in \cref{tab:main_results_150} with a total of $L=150$ private data parties controlling $M=500$ private samples in total. Other experimental settings are the same with those in \cref{tab:main_results_10}. Results show that, WASP continues to outperform baseline methods, including Pre-Text. This demonstrates that WASP is effectiveness not only under the need of guaranteeing sample-level DP but also under the need of providing user-level DP protection compared to baseline methods.

\begin{table*}[tb]
\caption{Evaluation of downstream STM accuracy using $L=150$. User-level DP is guaranteed instead of sample-level DP in this table. \textbf{Best} and \underline{second best} results are marked for each task (dataset, row).}
\label{tab:main_results_150}
\begin{center}
\begin{small}
\vspace{-1em}
    \begin{tabular}{c||c|c|cccccc|c}
    \toprule
        \multirow{2}{*}{~} & \multirow{2}{*}{\shortstack{\\Only\\Private}} & \multirow{2}{*}{\shortstack{\\\,\\FuseGen}} & \multicolumn{6}{c|}{Pre-Text} & \multirow{2}{*}{\shortstack{\\WASP\\(Ours)}} \\
        \cmidrule{4-9}
        ~ & ~ & ~ & GPT-2 & Llama-2 & Vicuna & OPT & ChatGLM3 & Flan-T5 & ~ \\
    \midrule
    \specialrule{0em}{0.3pt}{1pt}
    \midrule
        Privacy & $\epsilon=\infty$ & Aboluste Private & \multicolumn{6}{c|}{$(4.0,1\times10^{-5})$-DP} & $(4.0,1\times10^{-5})$-DP \\
        \midrule
        $|\mathcal{B}|$ & $500$ & - & \multicolumn{6}{c|}{$500$} & $500$ \\
        \midrule
        $|\mathcal{D}|$ & - & $6,000$ & \multicolumn{6}{c|}{$6,000$} & $6,000$\\
    \midrule
    \specialrule{0em}{0.3pt}{1pt}
    \midrule
        IMDb & 83.61 & \underline{89.09} & 83.96 & 84.28 & 83.67 & 84.69 & 85.56 & 88.71 & \textbf{89.15} \\
        Yelp-Rating & 44.15 & 57.96 & 45.78 & 50.54 & 51.42 & 50.40 & 51.54 & \underline{58.37} & \textbf{59.78} \\
    \bottomrule
    \end{tabular}
\end{small}    
\end{center}
\end{table*}

\begin{table*}[!htb]
\caption{Comparison of the information data parties' download, internal exchange and update in Pre-Text and WASP.}
\label{tab:upload_download_comparison}
\begin{small}
\begin{center}
\vspace{0.3em}
    \centering
    \begin{tabular}{c|ccc}
    \toprule
        ~ & Download & Exchange & Upload \\
    \midrule
        Pre-Text & embedding of each $(\mathbf{x}_i,y_i)\in\mathcal{D}$ & $\{H^{n}_l\}_{l=1}^L$ & $H^n$ \\
        WASP (Ours) & embedding of each $(\mathbf{x}_i,y_i)\in\mathcal{D}$ & $\{H^{n}_l\}_{l=1}^L,\{H^{f}_l\}_{l=1}^L$ & $H^n,H^f$ \\
    \bottomrule
    \end{tabular}
\end{center}
\end{small}
\end{table*}

\subsection{Comparison of Communication Overhead of WASP and Pre-Text for Federated Data Setting} \label{subsec:appendix_communication_overhead}

We compare the transmitted information for secure aggregation between the baseline method Pre-Text and our proposed WASP framework in \cref{tab:upload_download_comparison}. With the same number of participating data parties ($L$), WASP only requires aggregating additional $L$ histograms of dimension $\mathbb{R}^{|\mathcal{D}|}$ and uploading the aggregated histogram $H^f \in \mathbb{R}^{|\mathcal{D}|}$. These additional communicated information leads to only a minor increase in communication overhead compared to Pre-Text.

\section{Additional Related Works} \label{sec:appendix_related_work}
Due to space limitation, we include the discussion of previous works related to Contrastive In-context Learning (Contrastive ICL)
here in the Appendix.

\textbf{Contrastive In-context Learning.}\footnote{Works~\cite{ren2024towards,miyanishi2024multimodal} considering understanding in-context learning with contrastive learning theories are sometimes referred to using the same name, but we do not consider them here.}
The idea of using contrastive information to enrich in-context learning samples has been exploited from different aspects. Samples belonging to positive and negative classes~\cite{liang2024context}, correct or wrong self-predictions of training samples during training time~\cite{mo2024cicl}, human-preferred and non-preferred question responses~\citet{gao2024customizing} have all been utilized as contrastive samples.
Our study is the first known effort to consider contrastive in-context learning for synthetic data generation, by treating synthetic samples of different qualities generated by multiple PLMs as contrastive information.

\end{document}